%% file: 00-main-for-review.tex
\documentclass[10pt,twocolumn,letterpaper]{article}

\usepackage{iccv}
\usepackage{times}
\usepackage{epsfig}
\usepackage{graphicx}
\usepackage{amsmath}
\usepackage{amssymb}

\usepackage{bm}
\usepackage{bbm}
\usepackage{amsthm,amsmath,amssymb}
\usepackage{mathrsfs}
\usepackage{booktabs}
\usepackage{multirow}
\usepackage{bbding}
\usepackage{makecell}
\usepackage{enumitem}
\usepackage{caption}
\usepackage{subcaption}
\usepackage{graphicx}
\usepackage[detect-none]{siunitx}
\usepackage{wrapfig}
\usepackage{adjustbox}
\usepackage{color}
\usepackage{colortbl}
\usepackage[accsupp]{axessibility}

\definecolor{mygreen}{RGB}{46,139,87}
\definecolor{myblue}{RGB}{65,105,225}
\definecolor{lightgray}{RGB}{220,220,220}
\definecolor{darkgray}{RGB}{192,192,192}
\definecolor{mygray}{gray}{0.92}

\usepackage[table]{xcolor}

\newcommand{\hangjie}[1]{{\textcolor{black}{#1}}}

\usepackage{amssymb}% http://ctan.org/pkg/amssymb
\usepackage{pifont}% http://ctan.org/pkg/pifont
\newcommand{\xmark}{\ding{55}}%
\usepackage[pagebackref=true,breaklinks=true,letterpaper=true,colorlinks,bookmarks=false]{hyperref}
\usepackage[capitalize]{cleveref}
\crefname{section}{Sec.}{Secs.}
\Crefname{section}{Section}{Sections}
\Crefname{table}{Table}{Tables}
\crefname{table}{Tab.}{Tabs.}
\iccvfinalcopy % *** Uncomment this line for the final submission

\def\iccvPaperID{2669} % *** Enter the ICCV Paper ID here

\ificcvfinal\fi

\begin{document}

%%%%%%%%% TITLE
\title{RLIPv2: Fast Scaling of Relational Language-Image Pre-training}
% \title{RLIPv2: Fast Scaling of Relational Language-Image Pre-training \\ with Captioners}
% \title{RLIPv2: Fast Scaling of Relational Language-Image Pre-training \\ (with Captioners?)}

% \author{%
% Hangjie Yuan\textsuperscript{\rm 1,}\thanks{This work was done when they were interns at Alibaba DAMO Academy, supported by Alibaba Research Intern Program.} \hspace{.1in}
% Shiwei Zhang\textsuperscript{\rm 2} \hspace{.1in}
% Xiang Wang\textsuperscript{\rm 3,}\footnotemark[1] \hspace{.1in}
% Samuel Albanie\textsuperscript{\rm 4} \hspace{.1in}
% Yining Pan\textsuperscript{\rm 5,}\footnotemark[1] \hspace{.1in} \\
% Tao Feng\textsuperscript{\rm 2,} \hspace{.1in}
% Jianwen Jiang\textsuperscript{\rm 2} \hspace{.1in}
% Dong Ni\textsuperscript{\rm 1,}\thanks{Corresponding author.} \hspace{.1in}
% Yingya Zhang\textsuperscript{\rm 2}
% Deli Zhao\textsuperscript{\rm 2} \\
% \textsuperscript{\rm 1}Zhejiang University \hspace{.2in}
% \textsuperscript{\rm 2}Alibaba Group \hspace{.2in}
% \textsuperscript{\rm 3}Huazhong University of Science and Technology \\
% \textsuperscript{\rm 4}University of Cambridge \hspace{.2in}
% \textsuperscript{\rm 5}Singapore University of Technology and Design
% }

\vspace{-.3cm}
\author{%
    Hangjie Yuan\textsuperscript{\rm 1}\thanks{
    Work conducted during their research internships at DAMO Academy.} \hspace{.1in} 
    Shiwei Zhang\textsuperscript{\rm 2} \hspace{.1in} 
    Xiang Wang\textsuperscript{\rm 3}\footnotemark[1] \hspace{.1in} 
    Samuel Albanie\textsuperscript{\rm 4} \hspace{.1in} 
    Yining Pan\textsuperscript{\rm 5}\footnotemark[1] \hspace{.1in} \\
    Tao Feng\textsuperscript{\rm 2} \hspace{.1in}
    Jianwen Jiang\textsuperscript{\rm 2} \hspace{.1in}
    Dong Ni\textsuperscript{\rm 1}\thanks{Corresponding author.} \hspace{.1in} 
    Yingya Zhang\textsuperscript{\rm 2} \hspace{.1in} 
    Deli Zhao\textsuperscript{\rm 2} \vspace{0.2em} \\
    \vspace{-.1cm}
    \small
    \textsuperscript{\rm 1}Zhejiang University \hspace{.2in} 
    \textsuperscript{\rm 2}Alibaba Group \hspace{.2in} 
    \textsuperscript{\rm 3}Huazhong University of Science and Technology \\
    \small
    \textsuperscript{\rm 4}CAML Lab, University of Cambridge \hspace{.2in}
    \textsuperscript{\rm 5}Singapore University of Technology and Design 
    \\
    \small
    \texttt{\{hj.yuan, dni\}@zju.edu.cn} \hspace{.1in}  \texttt{wxiang@hust.edu.cn} \hspace{.1in}
    \texttt{\{pyn.sigrid, fengtao.hi, zhaodeli\}@gmail.com}
    \\
    \vspace{-.1cm}
    \small
    \texttt{sma71@cam.ac.uk} \hspace{.1in}
    \texttt{\{zhangjin.zsw, jianwen.jjw, yingya.zyy\}@alibaba-inc.com} 
    % \textsuperscript{\rm 1}Zhejiang University \hspace{.2in} 
    % \textsuperscript{\rm 2}Alibaba Group \hspace{.2in} 
    % \textsuperscript{\rm 3}Huazhong University of Science and Technology \\
    % \textsuperscript{\rm 4}University of Cambridge \hspace{.2in}
    % \textsuperscript{\rm 5}Singapore University of Technology and Design
    % \textsuperscript{\rm 4}National University of Singapore
    % \\ \texttt{\{hj.yuan, dni\}@zju.edu.cn} \hspace{.2in} \texttt{sma71@cam.ac.uk} \hspace{.2in} \texttt{ziyuan.huang@u.nus.edu}\\
    % \texttt{\{jianwen.jjw, shisi.ft, mingqian.tmq\}@alibaba-inc.com}
}

% \author{
%     Kunchang Li$^{1,2,3}$\thanks{Interns at Shanghai AI Laboratory. \textsuperscript{$\dag$}Corresponding authors.}\quad
%     Yali Wang$^{1,3}$\textsuperscript{$\dag$}\quad
%     Yizhuo Li$^{3,4}$\textsuperscript{*}\quad
%     Yi Wang$^{3}$\quad
%     Yinan He$^{3}$\\
%     Limin Wang$^{3,5}$\quad
%     Yu Qiao$^{1,3}$\textsuperscript{$\dag$}\vspace{0.2em}\\
%     % \textsuperscript{\dag}
%     \small{$^1$Shenzhen Institute of Advanced Technology, Chinese Academy of Sciences}\\
%     \small{$^2$University of Chinese Academy of Sciences\quad
%     $^3$Shanghai AI Laboratory\quad
%     $^4$The University of Hong Kong}\\
%     \small{$^5$State Key Laboratory for Novel Software Technology, Nanjing University}
% }

% \and

\maketitle
\ificcvfinal\thispagestyle{empty}\fi

%%%%%%%%% ABSTRACT
\begin{abstract}
    Relational Language-Image Pre-training (RLIP) aims to align vision representations with relational texts, thereby advancing the capability of relational reasoning in computer vision tasks.
    However, hindered by the slow convergence of RLIPv1\footnote{RLIPv1 refers to the model presented in ~\cite{Yuan2022RLIP}, and RLIPv2 refers to the model presented in this paper.} architecture and the limited availability of existing scene graph data, scaling RLIPv1 is challenging.
    In this paper, we propose RLIPv2, a fast converging model that enables the scaling of relational pre-training to large-scale pseudo-labelled scene graph data.
    To enable fast scaling, RLIPv2 introduces Asymmetric Language-Image Fusion (ALIF), a mechanism that facilitates earlier and deeper gated cross-modal fusion with sparsified language encoding layers.
    ALIF leads to comparable or better performance than RLIPv1 in a fraction of the time for pre-training and fine-tuning.
    To obtain scene graph data at scale, we extend object detection datasets with free-form relation labels by introducing a captioner (\textit{e.g.,} BLIP) and a designed Relation Tagger.
    The Relation Tagger assigns BLIP-generated relation texts to region pairs, thus enabling larger-scale relational pre-training.
    Through extensive experiments conducted on Human-Object Interaction Detection and Scene Graph Generation, RLIPv2 shows state-of-the-art performance on three benchmarks under fully-finetuning, few-shot and zero-shot settings.
    Notably, the largest RLIPv2 achieves 23.29mAP on HICO-DET without any fine-tuning, yields 32.22mAP with just 1\% data and yields 45.09mAP with 100\% data.
    Code and models are publicly available at \url{https://github.com/JacobYuan7/RLIPv2}.
   
\end{abstract}

%%%%%%%%% BODY TEXT
\input{01-intro}
\input{02-related-work}
\input{03-recap}
\input{04-methodology}
\input{05-experiments}
\input{06-conclusion}

{\small
\bibliographystyle{ieee_fullname}
\bibliography{egbib}
}

% \newpage
\clearpage
\input{08-appendix}

\end{document}

%% file: 01-intro.tex
\vspace{-.3cm}
\section{Introduction}
% \vspace{-.1cm}

\begin{figure}[t]
\centering
\includegraphics[width=0.46\textwidth]{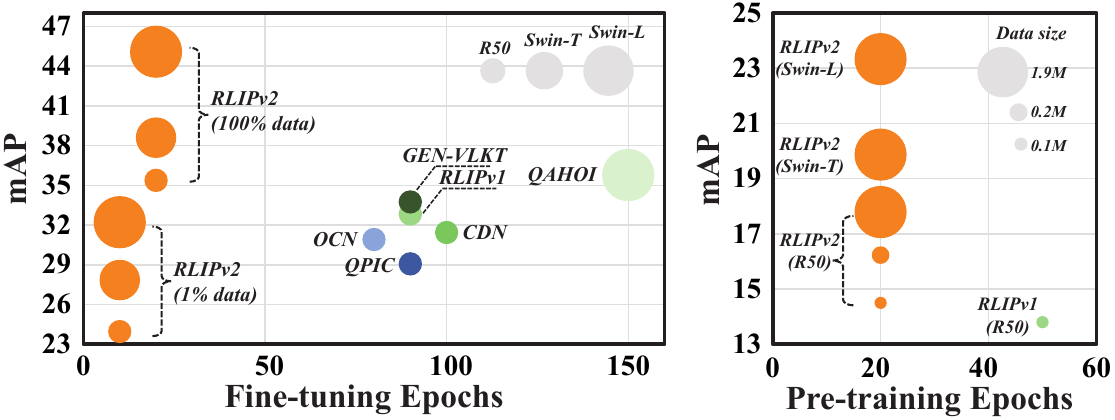}
\vspace{-0.1cm}
\caption{\small 
% \textbf{Left}: Fully-finetuning comparisons on HICO-DET.
\textbf{Left}: Fine-tuning comparison on HICO-DET.
\textbf{Right}: Pre-training epoch and zero-shot (NF) comparison on HICO-DET.
Except where stated, RLIPv2-ParSeDA architecture is adopted.
}
\vspace{-0.2cm}
\label{fig:performance_bubble}
\end{figure}

% The pretraining-finetuning paradigm has demonstrated its effectiveness in resolving various fundamental vision and language problems~\cite{liu2019RoBERTa,devlin2018BERT,he2019rethinking_imagenet,zoph2020rethinking_pretrain,radford2021CLIP}.
The pretraining-finetuning paradigm has achieved major breakthroughs in vision and language domains~\cite{liu2019RoBERTa,devlin2018BERT,he2019rethinking_imagenet,zoph2020rethinking_pretrain,radford2021CLIP,2023videocomposer}.
% With the recent emergence of large-scale Vision-Language Pre-training (VLP)~\cite{radford2021CLIP,li2022blip,li2023blip-2,wang2021simvlm,zhang2022glipv2,zeng2022x2-VLM}, the research community has witnessed the unprecedented scalability and generalization of an aligned pre-training.
In this context, a number of particularly notable results have been obtained through \textit{aligned} Vision-Language Pre-training (VLP)~\cite{radford2021CLIP,li2022blip,li2023blip-2,wang2021simvlm,zhang2022glipv2,zeng2022x2-VLM}.
% , the research community has witnessed the unprecedented scalability and generalization of an aligned pre-training.
% Such research projects typically rely on a robust \textbf{\textit{base model}}~\cite{dosovitskiy2020ViT_16_16,he2016resnet,dai2021dynamic_head,AttentionAlluNeed} that is trained on language-image paired \textbf{\textit{data}} to produce foundation models.
These research efforts have typically employed a robust \textbf{\textit{base model}}~\cite{dosovitskiy2020ViT_16_16,he2016resnet,dai2021dynamic_head,AttentionAlluNeed} that is trained on language-image paired \textbf{\textit{data}} to produce foundation models.
% In such research projects, a robust \textbf{\textit{base model}}~\cite{dosovitskiy2020ViT_16_16,he2016resnet,dai2021dynamic_head,AttentionAlluNeed} was chosen, fed with language-image paired \textbf{\textit{data}} and optimized to obtain foundation models.

% Various downstream tasks incorporate task-specific pre-training to scale-up performance .

RLIPv1~\cite{Yuan2022RLIP} presents the first attempt to specifically align vision representations and relational texts using VLP.
By pre-training on \textbf{\textit{open-vocabulary scene graph data}} like Visual Genome (VG)~\cite{krishna2017visualgenome}, RLIPv1 demonstrates its usefulness in zero-shot, few-shot and fully-finetuned Human-Object Interaction (HOI) Detection.
% Although RLIPv1 is proven effective, we find it intractable to be scaled up due to the following two limitations:
Although RLIPv1 is proven effective, we find it challenging to scale for the following reasons:

% \textbf{(i)} \textbf{\textit{Model}} perspective:
% Current RLIPv1 converges slowly, as exemplified by DETR~\cite{carion2020DETR}-based RLIPv1, which requires 150/90 epochs to converge during pre-training/fine-tuning. 
% Even when building on Deformable DETR (DDETR)~\cite{zhu2020deformableDETR} and DAB-DDETR~\cite{liu2022DABDETR}, 50/60 epochs are still required.

\textbf{(i)} \textbf{\textit{Model}} perspective:
RLIPv1 converges slowly, as exemplified by DETR~\cite{carion2020DETR}-based RLIPv1, which requires 150/90 epochs to converge during pre-training/fine-tuning. 
Even when building on Deformable DETR (DDETR)~\cite{zhu2020deformableDETR} and DAB-DDETR~\cite{liu2022DABDETR}, 50/60 epochs are still required.

\textbf{(ii)} \textbf{\textit{Data}} perspective:
as observed by the authors of RLIPv1, data with relation triplet annotations is scarce, impeding RLIPv1's scaling. 
Annotating triplets in the form of $\langle$\textit{subject, relations, object}$\rangle$ is both time- and cost-intensive.

To resolve the aforementioned challenges, we introduce RLIPv2, a fast converging model that enables relational pre-training on larger-scale pseudo-labelled scene graph data. 
% To resolve the above-mentioned challenges, we present RLIPv2, a fast scaling model to perform Relational Language-Image Pre-training on pseudo-labelled scene graph data. 
% on pseudo-labelled relational triplets. 
From both the model and data perspectives, we summarize the contributions of RLIPv2 as follows.

From the \textbf{\textit{model}} perspective, we observe that the slow convergence of DDETR can be attributed to the late language-image fusion strategy: fusion after decoding. 
Prior works~\cite{dou2022FIBER,li2021GLIP} have demonstrated an earlier and deeper fusion mechanism facilitates cross-modal alignment. 
In light of this, we propose Asymmetric Language-Image Fusion (ALIF) in RLIPv2 that encourages fusion in the encoding stage with sparsified language layers.
% In light of this, we propose Asymmetric Language-Image Fusion (ALIF) in RLIPv2 that encourages fusion in the encoding stage in an asymmetric architecture, enabling .
Without sacrificing inference speed thanks to sparsification, RLIPv2 requires only 20 epochs to pre-train and fine-tune based on DDETR family models~\cite{zhu2020deformableDETR,liu2022DABDETR} as shown in~\cref{fig:performance_bubble}, while performing better than or comparably to RLIPv1.

From the \textbf{\textit{data}} perspective, we leverage well-established object detection datasets~\cite{lin2014MSCOCO,shao2019Objects365,kuznetsova2020open_image}.
% prevailing and well-established object detection foundations
% , as shown in~\cref{fig:pseudo-labelling_pipeline}. 
Specifically, we extend these datasets with relational annotations by pseudo-labelling. 
To perform pseudo-labelling, we must tackle two challenges:
\textbf{(i)} sorting out the relations contained in the image and 
\textbf{(ii)} tagging relation texts to region pairs.
Regarding the first challenge, we employ external captioners (\textit{e.g.} BLIP~\cite{li2022blip}) that generate captions containing relation descriptions.
Regarding the second challenge, we reuse RLIPv2 model as a Relation Tagger (R-Tagger) that enables assigning the generated open-vocabulary relation texts to region pairs.
% By introducing a captioner (\textit{e.g.} BLIP~\cite{li2022blip}) and a designed Relation Tagger (R-Tagger), we tag open-vocabulary relation labels for exhaustive subject-object region pairs. 
% \textit{subject-object} 
Equipped with such a pipeline, we investigate the scaling behavior of both the model and the data for RLIPv2, which demonstrates improved zero-shot, few-shot and fine-tuning performance.

Furthermore, we introduce Scene Graph Generation (SGG), an analogously defined task to HOI detection for evaluating RLIPv2.
RLIPv2 achieves state-of-the-art performance on Open Images v6~\cite{kuznetsova2020open_image} for SGG, which underscores its robustness and efficacy in tackling relational reasoning tasks.

% "Furthermore, we introduce Scene Graph Generation (SGG), an analogously defined task to HOI detection for evaluating RLIPv2. RLIPv2 achieves state-of-the-art performance on Open Images v6~\cite{kuznetsova2020open_image} for SGG. Moreover, the superior results obtained by RLIPv2 on the Open Images v6 dataset underscore its robustness and efficacy in tackling Scene Graph Generation tasks."

% Other than HOI detection that RLIPv1 focuses on, we introduce Scene Graph Generation (SGG), an analogously defined task to HOI detection for evaluating RLIPv2. RLIPv2 shows state-of-the-art SGG performance on Open Images v6~\cite{kuznetsova2020open_image}.

%% file: 02-related-work.tex
\section{Related Work}
% 和scene graph文章的相似之处和区别
% 区别：我们用生成的，只用1个显然不好。
% 区别2：我们用了tagger而不是直接进行匹配。

\begin{figure*}[t]
\centering
\includegraphics[width=1.\textwidth]{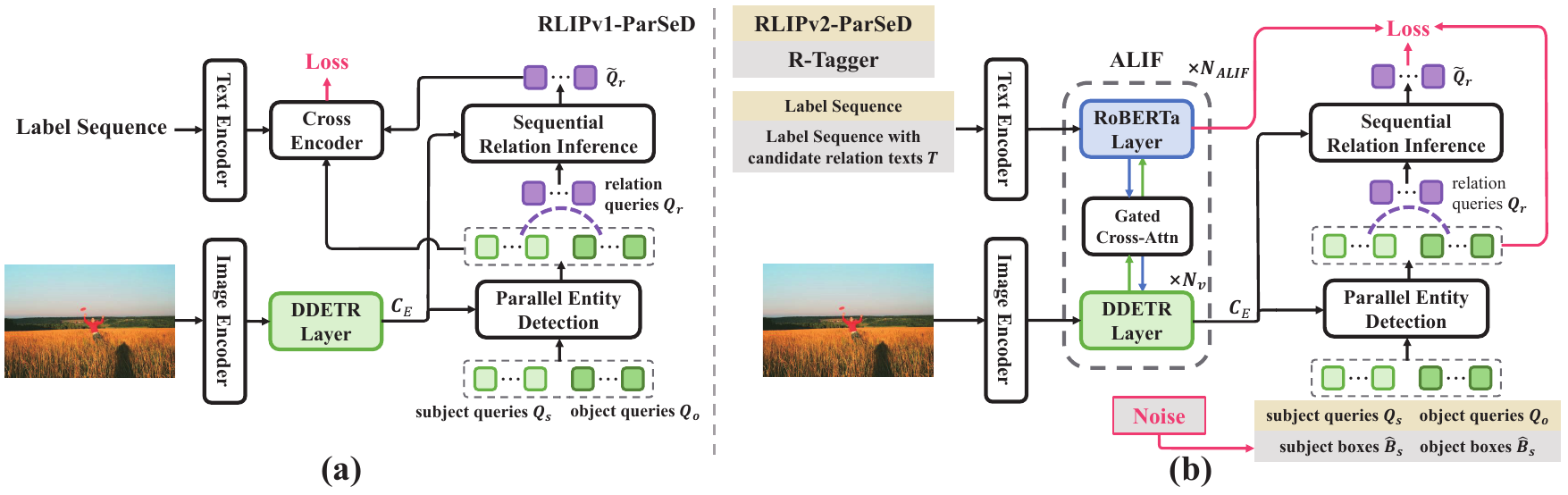}
\vspace{-0.4cm}
\caption{\small 
\textbf{The overview of (a) RLIPv1-ParSeD and (b) RLIPv2-ParSeD and R-Tagger.}
The red part (loss calculation and noise injection) is only valid during training.
In (a), \textit{Cross-Modal Fusion} is achieved by the cross encoder.
In (b), \textit{Cross-Modal Fusion} is achieved by ALIF.
The two architectures have an equivalent number of DDETR layers.
% The \textit{Parallel Entity Detection} and \textit{Sequential Relation Inference} blocks represent independent DETR-style decoders responsible for entity and relation detection, respectively.
% We omit the localisation loss for clarity.
}
\vspace{-.2cm}
\label{fig:RLIPv1_and_RLIPv2}
\end{figure*}

% \textbf{Language-image pre-training for vision tasks.}
% Recently, there are a series of representative works targeting learning visual representations from language supervision.
% Broadly, several works shows that including models for general purpose ~\cite{li2022blip,li2023blip-2,radford2021CLIP,alayrac2022flamingo,Jia2021ALIGN,kamath2021MDETR,dou2022FIBER,li2021GLIP,zhang2022glipv2}.
\textbf{Language-image pre-training for detection.}
% Recently, there are a series of representative works targeting learning visual representations from language supervision~\cite{li2022blip,li2023blip-2,radford2021CLIP,alayrac2022flamingo,Jia2021ALIGN,kamath2021MDETR,dou2022FIBER,li2021GLIP,zhang2022glipv2,wang2022beit-3}.
Recently, there has been a growing interest in learning visual representations from language supervision~\cite{li2022blip,li2023blip-2,radford2021CLIP,alayrac2022flamingo,Jia2021ALIGN,kamath2021MDETR,dou2022FIBER,li2021GLIP,zhang2022glipv2,wang2022beit-3}.
This paradigm of learning from language supervision has also proven effective in improving detection performance.
MDETR~\cite{kamath2021MDETR} was the first work to learn region-text correspondences in an end-to-end manner, while X-DETR~\cite{cai2022x-detr} improved upon MDETR by removing the cross-modal fusion part that improves its training efficiency.
GLIP~\cite{li2021GLIP} extended this line of research by scaling to web-scale data, leading to significant advances in zero-shot object detection and data efficiency.
DetCLIP~\cite{yao2022detclip} proposed a paralleled concept formulation and a concept dictionary to enable semantically rich region-text alignment.
RLIPv1~\cite{Yuan2022RLIP} was the first work to seek language-image alignment via relations, and our work follows its footsteps to achieve fast scaling of relational pre-training.
% It proposes Label Sequence Extension to enrich language concepts during contrastive learning, and Relation Quality Labels and Relation Pseudo-Labels to improve optimization using noisy samples.
% We follow RLIPv1 to fastly scale up its pre-training scale.

\textbf{End-to-end HOI detection and scene graph generation.}
Relations can interpret visual content in a fine-grained perspective~\cite{cong2023GICON,ji2020actiongenome,johnson2015imageRt}.
Detecting and recognizing relations utilizing HOI detection and scene graph generation have been verified effective in image captioning~\cite{yao2018exploringVR}, image retrieval~\cite{johnson2015imageRt,wang2020retrieval_scene_graph}, synthesis~\cite{yang2022diffusion_scene_graph,farshad2023scenegenie}, activity understanding~\cite{ji2020actiongenome,yuan2021DIN,yuan2021learningcontext} and .
The aim of these tasks is to detect relation triplets from a given input image.
Before the emergence of DETR~\cite{carion2020DETR}, the commonly adopted pipeline was to adopt an off-the-shelf object detector~\cite{ren2015faster,he2017mask} and design reasoning modules to infer relations~\cite{gao2018ican,gao2020DRG,Qi2018GPNN,li2019interactiveness,li2020hoianalysis,zellers2018neuralmotif,chen2019KERN,lin2020gps,li2021BGNN,li2018factorizable,xu2017iterative_message}.
Initial end-to-end design efforts extended detectors to support relation recognition~\cite{liao2020ppdm,zhong2021GGNet,yang2018graph_rcnn,kim2020uniondet,liu2021FCSGG}.
Further attempts focus on the adaptation of DETR to the field of HOI detection~\cite{Yuan2022OCN,chen2021ASNet,kim2021hotr,zou2021HOITransformer,tamura2021qpic,zhang2021CDN,zhang2022upt,wu2022cross-person_cues,kim2022mstr,liu2022interactiveness_field,zhou2022disentangled_transformer,park2022consistency_learning,qu2022distillation_using_oracle_queries,zhang2022STIP} and SGG~\cite{Li2021SGTR,cong2022RelTR}.
% RLIPv1~\cite{Yuan2022RLIP} was the first work to perform relational pre-training in boosting, demonstrating performance boost and data efficiency.
RLIPv1~\cite{Yuan2022RLIP} demonstrated the effectiveness of relational pre-training in improving performance and data efficiency. 
RLIPv2 builds upon RLIPv1, but seeks to solve its scaling problem.
The most related work~\cite{zhong2021scene_graph_language} also seeks scaling with the help of language. 
However, it remains a two-stage detection pipeline, lacks semantic richness and adopts a naive matching algorithm that hinders its performance.
% adopts a naive matching algorithm (R-Tagger). 

%% file: 03-recap.tex
\section{Recap of RLIPv1} \label{sec:recap}

%% In this section, we aim to briefly review RLIPv1. RLIPv1 is the first work to perform VLP that leverages both entity and relation descriptions.
%
% In this section, we will briefly review RLIPv1~\cite{Yuan2022RLIP}, the first work that leverages both entity and relation descriptions to perform VLP.
In this section, we will briefly review RLIPv1~\cite{Yuan2022RLIP}, a model that leverages both entity and relation descriptions to perform VLP and that forms the basis for our approach.
As triplet detection architecture, RLIPv1 proposes a ParSeD model that allocates decoupled embeddings for subjects, objects and relations. 
%
%% (Since RLIPv2 targets fast scaling, we only focus on DDETR-based detection architecture.)
%
%% The collective model RLIPv1-ParSeD is comprised of three stages: \textit{Parallel Entity Detection}, \textit{Sequential Relation Inference} and \textit{Cross-Modal Fusion}, which are outlined in~\ref{fig:RLIPv1_and_RLIPv2}(a).
%
The collective model RLIPv1-ParSeD consists of three stages: \textit{Parallel Entity Detection}, \textit{Sequential Relation Inference} and \textit{Cross-Modal Fusion}, as shown in~\cref{fig:RLIPv1_and_RLIPv2}(a).

For \textit{Parallel Entity Detection}, RLIPv1-ParSeD defines two sets of queries $\bm{Q}_{s}, \bm{Q}_{o} \in \mathbb{R}^{N_Q \times D}$ with $N_Q$ pairs of subjects and objects to perform subject and object detection.
For \textit{Sequential Relation Inference}, the model generates relation queries $\bm{Q}_{r} \in \mathbb{R}^{N_Q \times D}$ based on the decoded subject and object queries $\Tilde{\bm{Q}}_{s}, \Tilde{\bm{Q}}_{o} \in \mathbb{R}^{N_Q \times D}$, and performs decoding to obtain $\Tilde{\bm{Q}}_{r} \in \mathbb{R}^{N_Q \times D}$ for relation recognition.
The design of RLIPv1-ParSeD obeys the following  probabilistic factorization:
% \begin{equation}
% \mathbb{P}(\bm{G}|\bm{Q}_{s}, \bm{Q}_{o}, \bm{C};\bm{\theta}_{Par}, \bm{\theta}_{Se}) 
% = \mathbb{P}(\bm{B}_{s}, \bm{B}_{o}|\bm{Q}_{s}, \bm{Q}_{o}, \bm{C};\bm{\theta}_{Par})
% \cdot
% \mathbb{P}(\bm{R}|\bm{B}_{s}, \bm{B}_{o}, \bm{C};\bm{\theta}_{Se})
% \label{eqn:factorisation}
% \end{equation}
\begin{equation} \small
\begin{split}
&\mathbb{P}(\bm{G}|\bm{Q}_{s}, \bm{Q}_{o}, \bm{C}_{E}; \bm{\theta}_{Par}, \bm{\theta}_{Se}) 
= \\
&\mathbb{P}(\bm{B}_{s}, \bm{B}_{o}|\bm{Q}_{s}, \bm{Q}_{o}, \bm{C}_{E}; \bm{\theta}_{Par})
\cdot
\mathbb{P}(\bm{R}|\bm{B}_{s}, \bm{B}_{o}, \bm{C}_{E}; \bm{\theta}_{Se})
\label{eqn:factorisation}
\end{split}
\end{equation}
where $\bm{C}_{E}$ denotes features from the DDETR encoder{\footnote{Since RLIPv2 targets fast scaling, we only focus on DDETR-based detection architecture.};
$\bm{\theta}_{Par}, \bm{\theta}_{Se}$ denote parameters for \textit{Parallel Entity Detection} and \textit{Sequential Relation Inference};
$\bm{B}_{s}, \bm{B}_{o}, \bm{R}$ are sets of detected subject boxes, object boxes and relations, respectively.
They collectively comprise the detection results $\bm{G}$.

For \textit{Cross-Modal Fusion}, RLIPv1-ParSeD appends additional Transformer encoding layers~\cite{AttentionAlluNeed,kamath2021MDETR,li2021ALBEF} to perform language-image feature fusion on top of the decoded relation features $\Tilde{\bm{Q}}_{r}$, entity label features $\bm{L}_{E} \in \mathbb{R}^{N_E \times D}$ and relation label features $\bm{L}_{R} \in \mathbb{R}^{N_R \times D}$. 
$\bm{L}_{E}, \bm{L}_{R}$ are extracted from RoBERTa~\cite{liu2019RoBERTa}.

%% file: 04-methodology.tex
\section{Methodology}
%% In this section, we aim to introduce:
% \textit{i)} the architectural contribution Asymmetric Language-Image Fusion (ALIF) as the \textit{Cross-Modal Fusion} part in RLIPv2, as shown in~\ref{fig:RLIPv1_and_RLIPv2}(b);
% \textit{ii)} the overall framework of extending off-the-shelf object detection foundations~\cite{lin2014MSCOCO,shao2019Objects365} to support for RLIP. 
%
In this section, we will introduce:
\textbf{(i)} Asymmetric Language-Image Fusion (ALIF) as an efficient \textit{Cross-Modal Fusion} mechanism in RLIPv2, as shown in~\cref{fig:RLIPv1_and_RLIPv2}(b);
\textbf{(ii)} the overall framework of extending the off-the-shelf object detection datasets~\cite{lin2014MSCOCO,shao2019Objects365} to support relational pre-training.
% The framework includes the use of the BLIP captioner to perform image-level relation candidate set generation and the use of a designed Relation Tagger (R-Tagger) to tag relation descriptions for subject-object (SO) region pairs.

\subsection{Asymmetric Language-Image Fusion} \label{sec:ALIF}
The core idea underpinning ALIF is to perform efficient cross-modal fusion in the early stages of RLIPv2 as highlighted by~\cite{dou2022FIBER,li2021GLIP}.
Unlike RLIPv1 that encourages cross-modal alignment for entities and relations after the decoding phase, ALIF performs this during the detection encoding phase.
% ALIF encourages cross-modal alignment for entity and relation representations during the detection encoding phase rather than after the decoding phase. 
% \zsw{
% TODO: show the advantage and background knowledge to show why we do this. 
% How?
% }
%
This is particularly challenging for DDETR's encoder, since it relies on deformable attention that makes it challenging to adopt dedicated encoder layers during the detection encoding phase like~\cite{kamath2021MDETR,Maaz2021MViT,dou2022METER}. 

To address this, we propose ALIF, a mechanism that leverages DDETR encoding for the vision branch, RoBERTa encoding for the language branch and gated cross-attention for fusion.
% In ALIF, regarding the vision branch, ALIF performs DDETR encoding.
% Regarding the language branch, ALIF performs RoBERTa language encoding.
In contrast to previous work that encodes image and language with an equivalent number of layers~\cite{kamath2021MDETR,dou2022METER,li2021GLIP,dou2022FIBER}, we experimentally find that excessive RoBERTa layers do not improve its generalization capability due to its potential for overfitting to pre-trained data. 
Moreover, such a paradigm results in training difficulty due to the increased model complexity.
As a result, we perform DDETR encoding densely while performing RoBERTa encoding sparsely.
% empirically
We denote the vision features from the backbone as $\bm{C}^{(0)}$ and language features from RoBERTa as $\bm{L}^{(0)}$ (the concatenation of $\bm{L}_{E}$ and $\bm{L}_{R}$).
The first ALIF module can be formulated as:
\begin{equation}
    \Tilde{\bm{C}}^{(0)}, \Tilde{\bm{L}}^{(0)} = {\rm Cross\text{-}attn}(\bm{C}^{(0)}, \bm{L}^{(0)})
\end{equation}
\begin{equation}
    \bm{C}^{(N_{v})} = {\rm DDETR}^{N_{v}}(\bm{C}^{(0)} + {\rm G}(\Tilde{\bm{C}}^{(0)}))
\end{equation}
\begin{equation}
    \bm{L}^{(1)} = {\rm RoBERTa}^{1}(\bm{L}^{(0)} + {\rm G}(\Tilde{\bm{L}}^{(0)}))
\end{equation}
where $\Tilde{\bm{C}}^{(0)}$ denotes language features aggregated by cross attention, and $\Tilde{\bm{L}}^{(0)}$ is analogously defined; 
$N_{v}$ is the number of DDETR layers in one ALIF;
${\rm G}(\bm{x})$ defines a gating function that gates the aggregated cross-modal features.
Note that $\bm{C}_{E}$ is $\bm{C}^{(N_{v})}$ after stacking $N_{ALIF}$ ALIF layers for encoding.
For the instantiation of ${\rm G}(\bm{x})$, we experiment with three options:
\textbf{(i)} ${\rm G}(\bm{x}) = \alpha \bm{x}$ where $\alpha$ is a learnable scalar;
\textbf{(ii)} ${\rm G}(\bm{x}) = \bm{a} \bm{x}$ where $\bm{a}$ is a learnable vector;
\textbf{(iii)} ${\rm G}(\bm{x}) = {\rm SE}(\bm{x})$ where ${\rm SE}$ denotes a Squeeze-and-Excitation block~\cite{Hu2019SqueezeandExcitationN} with a reduction ratio of $4$.
We also experiment augmenting ${\rm G}(\bm{x})$ with ${\rm tanh}()$ introduced in~\cite{alayrac2022flamingo} (\textit{e.g.} ${\rm tanh}(\alpha \bm{x})$), while only observing side effects.
This indicates the magnitude of language and vision features are disparate, and a proper gating method is desired.

% We find deeper language models result in difficult training and high computation overhead.

\subsection{Relational Pseudo-labelling} \label{sec:relational_pseudo_labelling}
\begin{figure}[t]
\centering
\includegraphics[width=0.46\textwidth]{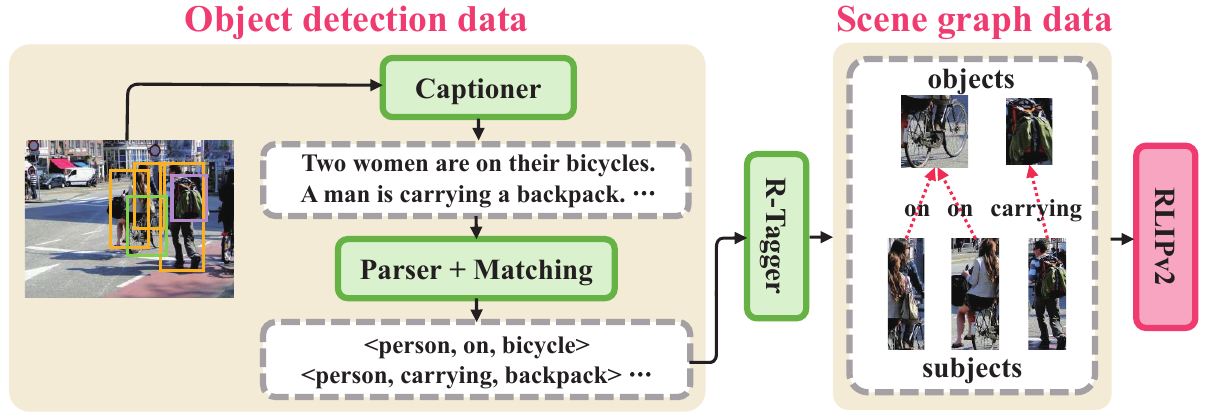}
\vspace{-0.1cm}
\caption{\small \textbf{An overview of relational pseudo-labeling}, which tags object detection data with free-form relation texts.
% This core of this process contains a captioner and a Relation Tagger (R-Tagger).
This process enables pre-training to be performed on two kinds of data.
}
\vspace{-0.2cm}
\label{fig:pseudo-labelling_pipeline}
\end{figure}

RLIPv2 reuses object detection datasets to benefit pre-training by pseudo-labelling, as shown in~\cref{fig:pseudo-labelling_pipeline}.
We assume that during the pre-training phase, the downstream entity and relation distributions are unavailable to us.
Then, to tag relation texts, we need to sort out the relations contained in the image as detailed in~\cref{sec:relation_candidate_set_generation}, and to tag relation texts to region pairs as detailed in~\cref{sec:relation_tagger}.
% 然后开始讲一下这个naive approach！！！！！！ 
% An intuitive approach is executing the pre-trained RLIPv2 model to perform SGG on object detection datasets (e.g. COCO and Objects365) to obtain pseudo triplets.
% However, this process bears two drawbacks:
% \textbf{(i)} specifying a suitable set of relation candidates for a given image is challenging;
% \textbf{(ii)} specifying SO region pairs that might have valid relations for a given image is challenging;
% \textbf{(iii)} during the execution of RLIPv2, the informative object annotations from object detection datasets fail to be utilized, thus degrading the quality of pseudo-labels.
% Aiming to solve these, we propose a framework that utilizes a BLIP captioner~\cite{li2022blip} to generate open-vocabulary relation candidate sets for images and a designed Relation Tagger (R-Tagger) to tag relations provided object annotations and relation candidate sets.

\vspace{-.2cm}
\subsubsection{Relation Candidate Set Generation} ~\label{sec:relation_candidate_set_generation}
% BLIP is able to generate free-form captions for a given image.
%% For a given image, relation candidate set generation aims to generate a coarse-grained set of candidate SO region pairs $\bm{P}$ and their candidate relation texts $\bm{T}$.
%
For a given image, relation candidate set generation aims to generate a coarse-grained set of candidate subject-object (SO) region pairs $\bm{P}$ and their candidate relation texts $\bm{T}$.
First,  we adopt BLIP to generate $N_{Cap}$ captions for each image.
Note that when $N_{Cap}=1$, we generate the caption via beam search, a deterministic generation method;
when $N_{Cap}>1$, we generate captions via nucleus sampling~\cite{holtzman2019nucleussampling}, a stochastic generation methods with cumulative probability threshold set to $0.9$.
Nucleus sampling adds semantic diversity to the generated captions, which contributes to diversity in the relations.

Second, we adopt a scene graph parser~\cite{schuster2015scene_graph_parser} to parse the obtained captions into relation triplets.
To filter out invalid parsed triplets, we perform string matching and keep those whose subjects and objects can be matched with any entities' names or entities' synonyms within the image~\cite{Yuan2022RLIP}.
This operation encloses a small set of possible SO region pairs $\bm{P}$ and possible relation texts $\bm{T}$ for the pairs (which are \textbf{inputs to R-Tagger}), without needing to traverse all possible pairs.
% This operation encloses possible subject-object pairs and possible relation texts for the pairs.

% \textcolor{red}{Can we delete it the following paragraph?}
% We also compare with the baseline of selecting the candidate set from VG annotations (select the relation texts if the subject and object texts are matched).
% However, the quality of such a selection method largely trails the proposed method, as shown in~\cref{tab:necessity_of_BLIP}.

\vspace{-.2cm}
\subsubsection{Relation Tagger via RLIPv2 Architecture} \label{sec:relation_tagger}
% \zsw{
% \subsubsection{Relation Tagger via RLIPv2} \label{sec:relation_tagger}
% }

% \begin{figure}[t]
% \centering
% \includegraphics[width=0.46\textwidth]{ICCV2023/figures/R-Tagger_framework.pdf}
% \caption{\small An overview of R-Tagger.
% % , which tags object detection data with free-form relation texts.
% The red part (loss calculation and noise injection) is only valid during training.
% }
% % \vspace{-0.4cm}
% \label{fig:R-Tagger_pipeline}
% \end{figure}

% The above generation process can generate false positive candidates.
The R-Tagger aims to assign candidate relation texts $\bm{T}$ to candidate SO region pairs $\bm{P}$ for a given image.
While one alternative is executing the pre-trained RLIPv2 model to perform SGG on object detection datasets to obtain pseudo-triplets,
the informative object annotations fail to be utilized directly during the execution of RLIPv2.
As a result, the quality of pseudo-labels is degraded.
A prior work~\cite{zhong2021scene_graph_language} adopts a coarse rule-based pseudo-labelling method that employed a greedy matching algorithm to randomly assign one relation text to an SO region pair as long as the relation texts' corresponding SO texts match with the SO region pair.
An ``overlap" prior was applied to further filter the triplets (\textit{i.e.}, a triplet is deemed valid only if the subject and object are overlapped).
This method, however, introduces excessive false positives, causing degraded pre-training.
Thus, we propose to \textbf{reuse the RLIPv2 architecture to perform relation prediction given ground-truth SO region pairs} as shown in Figure~\cref{fig:RLIPv1_and_RLIPv2}(b).
% The end-to-end optimized RLIPv2 is better than two-stage methods. (应该把这个观点加进去)

As outlined in~\cref{sec:recap}, RLIPv2 also utilizes subject queries $\bm{Q}_{s}$ and object queries $\bm{Q}_{o}$ as input.
To allow for the utilization of object annotations as the input, we propose to replace the detection queries $\bm{Q}_{s}$ and $\bm{Q}_{o}$ with object embeddings~\cite{li2022dn-detr}, which aims to encourage \textit{Parallel Entity Detection} to reconstruct object representations and \textit{Sequential Relation Inference} to recognize relations.
Compared with~\cref{eqn:factorisation}, the probabilistic factorisation of R-Tagger during inference can be reformulated as:
\begin{equation} \small
\begin{split}
& \mathbb{P}(\bm{R}|\hat{\bm{B}}_{s}, \hat{\bm{B}}_{o}, \bm{C}_{E}; \bm{\theta}_{Par}, \bm{\theta}_{Se}) 
= \\
& \mathbb{P}(\Tilde{\bm{B}}_{s}, \Tilde{\bm{B}}_{o}|\hat{\bm{B}}_{s}, \hat{\bm{B}}_{o}, \bm{C}_{E}; \bm{\theta}_{Par})
\cdot
\mathbb{P}(\bm{R}|\Tilde{\bm{B}}_{s}, \Tilde{\bm{B}}_{o}, \bm{C}_{E}; \bm{\theta}_{Se})
\label{eqn:inference_R-Tagger}
\end{split}
\end{equation}
where $\hat{\bm{B}}_{s}, \hat{\bm{B}}_{o}$ denote the ground-truth SO boxes from the region pair set $\bm{P}$ that include their positions and labels);
$\Tilde{\bm{B}}_{s}, \Tilde{\bm{B}}_{o}$ denote contextualized SO representations.
To allow for decoding, we embed $\hat{\bm{B}}_{s}, \hat{\bm{B}}_{o}$ to have an equivalent dimension with queries.
Specifically, we use MLPs to project positions and label text embeddings, and concatenate them along the channel dimension to obtain query-like input.

\textbf{Denoising training of R-Tagger.}
The training losses of R-Tagger are identical to RLIPv1:
\begin{equation} \label{eqn:overall_loss}
    \mathcal{L} = \lambda_{1} \mathcal{L}_{l1} + \lambda_{2} \mathcal{L}_{GIoU} + \lambda_{3} (\mathcal{L}_{s} + \mathcal{L}_{o}) + \lambda_{4} \mathcal{L}_{r}
\end{equation}
where $\mathcal{L}$ is comprised of the $\ell_1$ loss for box regression $\mathcal{L}_{l1}$, GIoU loss~\cite{rezatofighi2019GIoU} $\mathcal{L}_{GIoU}$, Cross-Entropy (CE) loss for subject and object classes $\mathcal{L}_{s}, \mathcal{L}_{o}$, and Focal loss~\cite{lin2017focal} for relations $\mathcal{L}_{r}$.
$\lambda_{1}, \lambda_{2}, \lambda_{3}, \lambda_{4}$ are set to $2.5, 1, 1, 1$ as fixed weights to balance multi-task training~\cite{Yuan2022RLIP,Yuan2022OCN,tamura2021qpic}.

%% However, such a training procedure would encounter collapse since the input and training objectives of \textit{Parallel Entity Detection} are identical.
%% The model can be optimized to find a shortcut (identity mapping) to obtain the minimum loss value, thus causing model collapse.
However, due to the identical input and training objectives of \textit{Parallel Entity Detection}, the model will attempt to find a shortcut, \emph{i.e.}, identity mapping, to achieve the minimum loss value.
To avoid this, we draw inspiration from~\cite{li2022dn-detr} which adds noise to $\hat{\bm{B}}_{s}, \hat{\bm{B}}_{o}$ during training.
Specifically, we follow~\cite{li2022dn-detr} to add center shifting and box scaling noise to box positions and add label flipping noise to box labels.
The noise scale of center shifting and box scaling is set to $0.4$, and the noise scale of label flipping is set to $0.2$ following~\cite{li2022dn-detr}.
% Should we detail how the noise is added? And what is the meaning of the noise scale?
Furthermore, to prevent the information leakage between the same region with different noise, we apply attention masks to block the information flow between the same regions in \textit{Parallel Entity Detection}.

\textbf{Inference of R-Tagger.}
After the training of R-Tagger, we can use it to infer relations without additional noise based on~\cref{eqn:inference_R-Tagger}.
For each inference, the maximum number of region pairs is $N_{Q}$.
If candidate SO region pairs exceed $N_{Q}$, we infer multiple times and merge the results.
% The confidence of a relation is defined as the multiplication of the top-1 score from the softmax distribution over the subject, the object and the original relation ${\rm sigmoid}$ score.
To quantify the confidence of a relation, we calculate the product of the top-1 score from the ${\rm softmax}$ distribution over the subject, the object and the original relation ${\rm sigmoid}$ score.
To select pseudo-labels, we choose those whose relation confidence exceeds a threshold $\eta$.
$\eta$ is set to $0.2$ by default 
(details can be found in the Appendix).

% relation texts $\bm{T}$ to SO region pairs $\bm{P}$
% 需要设定threshold来过滤。
% 像query一样，也是多query来推理
% Relation score 是什么？三个乘一乘

% Verify the relation texts (not all matched triplets are valid.)

% \subsection{Pre-training, Fine-tuning and Inference of RLIPv2}
\subsection{Pre-training, Fine-tuning and Inference}
Regarding pre-training and fine-tuning of RLIPv2, we first merge results from \textit{Parallel Entity Detection} and \textit{Sequential Relation Inference} to obtain $N_{Q}$ triplets.
Next, we employ the bipartite matching algorithm originally proposed in~\cite{tamura2021qpic} to match the predicted and ground-truth triplet annotations.
% Provided triplet annotations, we use bipartite matching originally proposed in~\cite{tamura2021qpic} to match the ground-truth triplets and predicted ones.
The overall loss is identical to~\cref{eqn:overall_loss}.
The techniques introduced in RLIPv1, \textit{i.e.} Label Sequence Extension, Relation Quality Labels and Relation Pseudo-Labels, are employed by default for a fair comparison.
% The definition of the relation confidence is identical to the definition in~\cref{sec:relation_tagger}.
Regarding the inference of RLIPv2, we sort the relation confidence (defined in~\cref{sec:relation_tagger}) of the correctly localised triplets (${\rm IoU}>0.5$) and select the Top-$K$ triplets.
$K$ is set to 100 by default following~\cite{Yuan2022RLIP,tamura2021qpic,Li2021SGTR,cong2022RelTR}.
% bipartite matching
% loss (引用前面)
% RQL RPL and LSE
% 学一下RLIPv1的文章的写法。

%% file: 05-experiments.tex
\section{Experiments} \label{sec:experiments}

\begin{table}[t]
  % \small
  \setlength{\tabcolsep}{3pt}
  \centering
    \begin{tabular}{cc|ccc|cc}
    \toprule
    $N_{v}$ & $N_{ALIF}$ & \textbf{Rare} & \textbf{Non-Rare} & \textbf{Full} & \textbf{\#Params} & \textbf{FPS} \\
    \midrule
    \midrule
    1     & 6     & 10.92  & 13.99  & 13.28  & 246.8M & 18.93 \\
    2     & 3     & \textbf{12.12} & \textbf{14.07} & \textbf{13.62} & 206.6M & 21.49 \\
    3     & 2     & 10.57  & 14.05  & 13.25  & 193.2M & 22.15 \\
    6     & 1     & 11.26  & 13.90  & 13.30  & 179.8M & 23.17 \\
    \bottomrule
    \end{tabular}
    \vspace{-.1cm}
    \caption{\small \textbf{The effect of the sparsification of language encoding layers in \textit{Cross-Modal Fusion}}. Results are evaluated on HICO-DET under zero-shot (NF) setting. FPS (frames per second) is tested on a single NVIDIA A100 with a minibatch size 1.}
    \vspace{-.2cm}
  \label{tab:sparsification_language_encoder}
\end{table}%

\textbf{Pre-training datasets.}
To pre-train RLIPv2, we utilize VG~\cite{krishna2017visualgenome}, COCO~\cite{lin2014MSCOCO} and Objects365~\cite{shao2019Objects365}.
VG has 108$k$ images annotated with free-form relation and object annotations.
COCO has 117$k$ images with only object annotations in 80 classes, and Objects365 has 1,742$k$ images with only object annotations in 365 classes.
Thus, we use relational pseudo-labelling to tag relation labels for COCO and Objects365, enabling them to support relational pre-training.
% Thus, we use relational pseudo-labelling introduced in~\cref{sec:relational_pseudo_labelling} to tag pseudo relation labels for COCO and Objects365, enabling them to support relational pre-training.
% Thus, we use relational pseudo-labelling introduced in~\cref{sec:relational_pseudo_labelling} to tag pseudo relation labels for COCO and Objects365, enabling them to support relational pre-training.
% exclude 

% For HOI detection, we evaluate on the widely-used HICO-DET~\cite{chao2015hico} and V-COCO~\cite{gupta2015VisualSemanticRole} dataset.
% HICO-DET consists of 37,536 training images and 9,515 testing images, labelled
% with 600 HOI triplets derived from combinations of 117 verbs and 80 objects.
% We assess RLIPv2 on HICO-DET under the \textbf{Default} setting on \textit{Full}, \textit{Rare} and \textit{Non-Rare} sets.
% V-COCO consists of 2,533 training images, 2,876 validation images and 4,946 testing images, labelled with 24 interactions and 80 objects.
% As introduced in the official evaluation code~\cite{gupta2015VisualSemanticRole}, we evaluate under two scenarios:  ${\rm AP}_{role}^{\#1}$ and ${\rm AP}_{role}^{\#2}$.

\textbf{Downstream datasets.} 
For \textbf{HOI detection}, we follow~\cite{tamura2021qpic,Yuan2022RLIP,Yuan2022OCN,zhang2021CDN} to evaluate on HICO-DET~\cite{chao2015hico} and V-COCO~\cite{gupta2015VisualSemanticRole}.
For HICO-DET that contains 117 verbs and 80 objects, we evaluate under the \textbf{Default} setting on \textit{Full}, \textit{Rare} and \textit{Non-Rare} sets.
For V-COCO that contains 24 interactions and 80 objects, we evaluate following the official evaluation code~\cite{gupta2015VisualSemanticRole} under two scenarios:  ${\rm AP}_{role}^{\#1}$ and ${\rm AP}_{role}^{\#2}$.
For \textbf{SGG}, we assess RLIPv2 on the widely-used Open Images v6~\cite{kuznetsova2020open_image} dataset, which is annotated with 288 objects and 30 relations.
We evaluate RLIPv2 using standard evaluation metrics~\cite{kuznetsova2020open_image,zhang2019RelDN}: 
Recall@50 (${\rm R}@50$), weighted mean Average Precision for relation detection (${\rm wmAP}_{rel}$) and phrase detection (${\rm wmAP}_{phr}$).
The final score is calculated as:
${\rm score}_{wtd} = 0.2*{\rm R}@50+0.4*{\rm wmAP}_{rel}+0.4*{\rm wmAP}_{phr}$.

% For SGG, we assess RLIPv2 on the widely-used Open Images v6~\cite{kuznetsova2020open_image} dataset.
% Open Images v6 contains 126$k$ training images, 5.3$k$ test images and 1.8$k$ validation images, annotated with 288 objects and 30 relations.
% We evaluate RLIPv2 using standard evaluation metrics~\cite{kuznetsova2020open_image,zhang2019RelDN}: 
% Recall@50 (${\rm R}@50$), weighted mean Average Precision for relation detection (${\rm wmAP}_{rel}$) and phrase detection (${\rm wmAP}_{phr}$).
% The final score is calculated as:
% ${\rm score}_{wtd} = 0.2*{\rm R}@50+0.4*{\rm wmAP}_{rel}+0.4*{\rm wmAP}_{phr}$.

\textbf{Implementation details.}
To assess the effectiveness of RLIPv2, we choose to adopt DDETR~\cite{zhu2020deformableDETR} to compose \textbf{RLIPv2-ParSeD} and adopt DAB-DDETR~\cite{liu2022DABDETR} to compose \textbf{RLIPv2-ParSeDA}.
To peform a fair comparison with previous works, we ensure that RLIPv2 has $N_{v}*N_{ALIF}=6$ DDETR/DAB-DDETR encoding layers.
For \textit{Parallel Entity Detection} and \textit{Sequential Relation Inference}, 3 layers are adopted following~\cite{Yuan2022RLIP,Yuan2022OCN,tamura2021qpic,zhang2021CDN}.
$N_{Q}$ is set to 100 during pre-training and fine-tuning, except when we fine-tune on HICO-DET where $N_{Q}$ is set to 64 following~\cite{zhang2021CDN}.
Regarding model initialization, we use COCO detection parameters as initialisation when using VG or VG and COCO for pre-training; when using VG, COCO and Objects365 for pre-training, we use COCO and Object365 detection parameters as initialisation.
Regarding the configuration of minibatch sizes and learning rate (LR), we set the minibatch size to 64, LR for the text encoder to 1.41e-5 and LR for other modules to 1.41e-4 when using ResNet-50~\cite{he2016resnet} and Swin-T~\cite{liu2021swin};
We set the minibatch size to 32, LR for the text encoder to 1e-5 and LR for other modules to 1e-4 when using Swin-L~\cite{liu2021swin}.
RLIPv2 and R-Tagger are pre-trained and fine-tuned for 20 epochs unless otherwise stated, with LR dropping by a factor of 10 at the 15$th$ epoch.
For the BLIP captioner, we adopt the ViT-L/16~\cite{dosovitskiy2020ViT_16_16} model fine-tuned on COCO Caption~\cite{chen2015COCO_Caption}.

% \textbf{Experimental protocols.}
% \textit{Full}
% \textit{Rare}
% \textit{Non-Rare} set.

{\renewcommand{\arraystretch}{0.9}
\begin{table}[t]
  \small
  \setlength{\tabcolsep}{11pt}
  \centering
    \begin{tabular}{cc|ccc}
    \toprule
    \textbf{Gating} & {\rm tanh()} & \textbf{Rare} & \textbf{Non-Rare} & \textbf{Full}  \\
    \midrule    
    \midrule
    $\alpha$     & \multirow{3}[2]{*}{\Checkmark} & 10.47  & 13.83  & 13.05  \\
    $\bm{a}$ &       & 9.91  & 13.54  & 12.70  \\
    SE    &       & 11.07  & 13.89  & 13.24  \\
    \midrule
    $\alpha$     & \multirow{3}[2]{*}{\xmark} & \textbf{12.12} & \textbf{14.07} & \textbf{13.62} \\
    $\bm{a}$ &       & 10.98  & \textbf{14.07} & 13.36  \\
    SE    &       & 11.07  & 14.00  & 13.32  \\
    \bottomrule
    \end{tabular}%
    \vspace{-.1cm}
    \caption{\small \textbf{Comparisons of different gating functions} introduced in~\cref{sec:ALIF}. We report zero-shot (NF) results on HICO-DET.}
    \vspace{-.2cm}
  \label{tab:gating_function}%
\end{table}}

\begin{figure*}[t]
{\renewcommand{\arraystretch}{0.9}
\begin{minipage}[l]{0.63\textwidth}
  \small
  % \scriptsize
  \setlength{\tabcolsep}{2pt}
  \centering
    \begin{tabular}{c|ccc|cc|c}
    \toprule
    \multirow{2}[2]{*}{\textbf{Model}} & \multicolumn{3}{c|}{\textbf{Pre-training}} & \multicolumn{2}{c|}{\textbf{Fully-finetuning}} & \multirow{2}[2]{*}{\textbf{FPS}} \\
          % & \textbf{Epochs} & \textbf{Time (h)} & \textbf{Zero-shot (NF)} & \textbf{Epochs} & \textbf{Result} &  \\
          & \textbf{Ep.} & \textbf{Time} & \textbf{Zero-shot (NF)} & \textbf{Ep.} & \textbf{Result} &  \\
    \midrule
    RLIPv1-ParSeD & 50    & 25.9h  & 11.20 / \textbf{14.73} / \textbf{13.92} & 60    & 24.67 / 32.50 / 30.70 & 21.89 \\
    \rowcolor{mygray}  RLIPv2-ParSeD & 20    & 10.9h  & \textbf{12.12} / 14.07 / 13.62 & 20    & \textbf{26.47} / \textbf{33.51} / \textbf{31.89} & 21.49 \\
    \midrule
    RLIPv1-ParSeDA & 50    & 27.2h  & 11.34 / 14.56 / 13.82 & 60    & 22.85 / 30.87 / 29.03 & 19.41 \\
    \rowcolor{mygray}  RLIPv2-ParSeDA & 20    & 11.3h  & \textbf{13.03} / \textbf{14.98} / \textbf{14.53} & 20    & \textbf{27.01} / \textbf{35.21} / \textbf{33.32} & 18.94 \\
    \bottomrule
    \end{tabular}
    \vspace{-.1cm}
    \makeatletter\def\@captype{table}\makeatother\caption{\small \textbf{Comparisons of RLIPv1 and RLIPv2} under zero-shot (NF) and fully-finetuning settings on HICO-DET pre-trained on VG. Results are reported on \textit{Rare}/\textit{Non-Rare}/\textit{Full} sets. FPS is tested on a single NVIDIA A100 with minibatch size 1. Pre-training time is tested on 8 NVIDIA A100. ``Ep.'' denotes number of epochs. }
    \vspace{-.2cm}
  \label{tab:RLIPv1_and_RLIPv2}%
\end{minipage}}
\hspace{2pt}
{\renewcommand{\arraystretch}{0.9}
\begin{minipage}[r]{0.35\textwidth}
  % \small
  % \scriptsize
  \setlength{\tabcolsep}{3pt}
  \centering
    \begin{tabular}{c|ccc}
    \toprule
    \textbf{Initialization} & \textbf{Rare} & \textbf{Non-Rare} & \textbf{Full} \\
    \midrule
    \midrule
    COCO (default) & 12.12  & 14.07  & 13.62  \\
    \midrule
    R-Tagger & 12.55 & 13.64 & 13.39 \\
    % R-Tagger w/o noise & 9.94  & 12.74 & 12.09 \\
    % R-Tagger w/o attn masks & 8.55  & 11.22 & 10.61 \\
    w/o noise & 9.94  & 12.74 & 12.09 \\
    w/o attn masks & 8.55  & 11.22 & 10.61 \\
    \bottomrule
    \end{tabular}
    \vspace{-.1cm}
    \makeatletter\def\@captype{table}\makeatother\caption{\small \textbf{The quality of R-Tagger parameters.} Results are evaluated using RLIPv2-ParSeD with ResNet-50 under zero-shot (NF) setting. }
    \vspace{-.2cm}
  \label{tab:R-Tagger_params}
\end{minipage}}
\end{figure*}

\subsection{Ablation Study}
We perform the ablation study using RLIPv2-ParSeD with ResNet-50 as the backbone, and VG as the pre-training dataset unless otherwise stated.
We pre-train for 20 epochs and evaluate performance on HICO-DET under the zero-shot with no fine-tuning (NF) setting.

\vspace{-.2cm}
\subsubsection{Asymmetric Language-Image Fusion}
\textbf{Sparsification of the language layers.} 
First of all, we ablate on the number of language layers to investigate the effect of sparsifying language encoding layers in \textit{Cross-Modal Fusion}.
As introduced in~\cref{sec:ALIF}, dense language layers cause difficult training.
To avoid training collapse, we compute classification losses ($\mathcal{L}_{s}, \mathcal{L}_{o}$ and $\mathcal{L}_{r}$) using all intermediate language features rather than using only the last layer of language features in \textit{Cross-Modal Fusion}.
Results are shown in~\cref{tab:sparsification_language_encoder}.
Our findings suggest that sparsifying the language encoder does not compromise zero-shot performance and improves parameter efficiency.
We attribute this to overfitting effects on upstream datasets impairing models' generalization.
We choose $N_{v}=2$ and $N_{ALIF}=3$ as the default hyper-parameters for the following experiments.

% downgrade to no-fusion models.

\textbf{The choice of the gating function ${\rm G}(\bm{x})$.}
% To properly fuse cross-modal features, we experiment on the variants of the gating function as introduced in~\cref{sec:ALIF}.
To investigate the effect of the gating function on cross-modal fusion, we experiment with variants of the gating function as described in~\cref{sec:ALIF}.
The results are shown in~\cref{tab:gating_function}.
We note that ${\rm tanh}{}$ consistently degrades performance for the three gating methods.
We conjecture that the output range of ${\rm tanh}{}$ can constrain the magnitude of the features to be fused, thus limiting the performance.
Based on the results, we choose the simplest one: the learnable scalar gating method parameterised by $\alpha$.

\textbf{Comparisons of RLIPv2 with RLIPv1.}
To compare the architectural benefit of RLIPv2 with RLIPv1, we adopt DDETR~\cite{zhu2020deformableDETR} and DAB-DDETR~\cite{liu2022DABDETR} to follow two designs, and evaluate performance under zero-shot (NF) and fully-finetuning setting. Results are shown in~\cref{tab:RLIPv1_and_RLIPv2}.
From this table, we can observe that by modifying the architecture without compromising much inference speed, RLIPv2 generally outperforms its RLIPv1 counterpart.
Specifically, RLIPv2 obtains comparable zero-shot results to RLIPv1, while costing about 0.4$\times$ pre-training time.
In terms of fine-tuning results, RLIPv2 surpasses RLIPv1 with 0.33$\times$ fine-tuning time due to earlier and deeper fusion.

% \begin{table}[t]
%   % \small
%   \scriptsize
%   \setlength{\tabcolsep}{2pt}
%   \centering
%     \begin{tabular}{c|ccc|cc|c}
%     \toprule
%     % \textbf{Model} & \shortstack{\textbf{PT} \\ \textbf{Epochs}}& \shortstack{\textbf{PT} \\ \textbf{Time (h)}} & \textbf{Zero-shot (NF)} & \textbf{FFT Epochs} & \textbf{FFT} & \textbf{FPS} \\
%     \multirow{2}[2]{*}{\textbf{Model}} & \multicolumn{3}{c|}{\textbf{Pre-training}} & \multicolumn{2}{c|}{\textbf{Fully-finetuning}} & \multirow{2}[2]{*}{\textbf{FPS}} \\
%           & \textbf{Epochs} & \textbf{Time (h)} & \textbf{Zero-shot (NF)} & \textbf{Epochs} & \textbf{Result} &  \\
%     \midrule
%     \midrule
%     RLIPv1-ParSeD & 50    & 25.9  & 11.20 / \textbf{14.73} / \textbf{13.92} & 60    & 24.67 / 32.50 / 30.70 & 21.89 \\
%     RLIPv2-ParSeD & 20    & 10.9  & \textbf{12.12} / 14.07 / 13.62 & 20    & \textbf{26.47} / \textbf{33.51} / \textbf{31.89} & 21.49 \\
%     \midrule
%     RLIPv1-ParSeDA & 50    & 27.2  & 11.34 / 14.56 / 13.82 & 60    & 22.85 / 30.87 / 29.03 & 19.41 \\
%     RLIPv2-ParSeDA & 20    & 11.3  & \textbf{13.03} / \textbf{14.98} / \textbf{14.53} & 20    & \textbf{27.01} / \textbf{35.21} / \textbf{33.32} & 18.94 \\
%     \bottomrule
%     \end{tabular}
%     \caption{\small Comparisons of RLIPv1 and RLIPv2 under zero-shot (NF) and fully-finetuning settings on HICO-DET pre-trained on VG. Results are reported on \textit{Rare}/\textit{Non-Rare}/\textit{Full} sets. FPS is tested on a single NVIDIA A100 with minibatch size 1. Pre-training time is tested on 8 NVIDIA A100.}
%   \label{tab:RLIPv1_and_RLIPv2}%
% \end{table}%

\vspace{-.2cm}
\subsubsection{Relational Pseudo-labelling} \label{sec:exp_relational_pseudo_labelling}
% The training of R-Tagger is detailed in~\cref{sec:experiments}.
By default, we use RLIPv2-ParSeD with ResNet-50 backbone as the basic structure of R-Tagger.

\textbf{The necessity of denoising pre-training and attention masks for R-Tagger.}
If noise is not added during R-Tagger's pre-training, the loss will fluctuate instead of steadily decreasing as the optimization proceeds.
If attention masks are not utilized to prevent information leakage, the model will tend to learn an identity mapping, thus degrading its ability to infer relations.
R-Tagger is pre-trained for 20 epochs.
To assess the quality of R-Tagger's parameters, thanks to R-Tagger's identical structure to RLIPv2-ParSeD, we initialize RLIPv2-ParSeD with R-Tagger's parameters and pre-train for 10 epochs.
We evaluate the zero-shot (NF) performance on HICO-DET as shown in~\cref{tab:R-Tagger_params}.
We observe that removing additional noise or attention masks both impair performance, highlighting their importance.

% \begin{table}[t]
%   % \small
%   \scriptsize
%   \setlength{\tabcolsep}{3pt}
%   \centering
%     \begin{tabular}{c|ccc}
%     \toprule
%     \textbf{Initialization} & \textbf{Rare} & \textbf{Non-Rare} & \textbf{Full} \\
%     \midrule
%     \midrule
%     COCO (default) & 12.12  & 14.07  & 13.62  \\
%     \midrule
%     R-Tagger & 12.55 & 13.64 & 13.39 \\
%     R-Tagger w/o noise & 9.94  & 12.74 & 12.09 \\
%     R-Tagger w/o attn masks & 8.55  & 11.22 & 10.61 \\
%     \bottomrule
%     \end{tabular}%
%     \caption{\small The quality of R-Tagger parameters. Results are evaluated using RLIPv2-ParSeD with ResNet-50 under zero-shot (NF) setting.}
%   \label{tab:R-Tagger_params}
% \end{table}

{\renewcommand{\arraystretch}{0.9}
\begin{table}[t]
  \small
  \setlength{\tabcolsep}{4.5pt}
  \centering
    \begin{tabular}{cc|ccc}
    \toprule
    \textbf{Tagging Method} & \textbf{Overlap} & \textbf{Rare} & \textbf{Non-Rare} & \textbf{Full} \\
    \midrule
    \midrule
    \multirow{2}[2]{*}{Greedy~\cite{zhong2021scene_graph_language}} & \xmark & 11.15  & 11.65  & 11.55  \\
          & \Checkmark & 13.16  & 14.70  & 14.35  \\
    \midrule
    \multirow{2}[2]{*}{CLIP (ViT-L/14)~\cite{radford2021CLIP}} & \xmark & 12.66  & 12.76  & 12.74  \\
          & \Checkmark     & 14.63  & 14.94  & 14.87  \\
    \midrule
    \multirow{2}[2]{*}{R-Tagger (ResNet-50)} & \xmark & 15.33  & \textbf{15.54}  & \textbf{15.49}  \\
          & \Checkmark & \textbf{15.36}  & 15.37  & 15.36  \\
    \bottomrule
    \end{tabular}%
    \vspace{-.1cm}
    \caption{\small \textbf{Comparisons of relation tagging methods.} ``Overlap'' denotes the ``overlap'' prior for SO pairs introduced in~\cref{sec:relation_tagger}. We report zero-shot (NF) results pre-trained on VG and COCO. We use oracle captions from \textbf{COCO Caption}~\cite{chen2015COCO_Caption} ($N_{Cap}=5$).}
    \vspace{-.2cm}
  \label{tab:tagging_strategy}
\end{table}}

{\renewcommand{\arraystretch}{0.9}
\begin{table}[t]
  \small
  \setlength{\tabcolsep}{7.5pt}
  \centering
    \begin{tabular}{cc|ccc}
    \toprule
    \textbf{Caption type} & $N_{Cap}$ & \textbf{Rare} & \textbf{Non-Rare} & \textbf{Full} \\
    \midrule
    \midrule
    Oracle & 5     & 15.33  & 15.54  & 15.49  \\
    \midrule
    BLIP (beam) & 1     & 9.86  & 12.02  & 11.52  \\
    BLIP (nucleus) & 5     & 14.67  & 14.76  & 14.74  \\
    BLIP (nucleus) & 10    & \textbf{15.08} & \textbf{15.10} & \textbf{15.09} \\
    BLIP (nucleus) & 20    & 14.24  & 14.91  & 14.75  \\
    \midrule
    BLIP (nucleus)\textsuperscript{*} & 5     & 12.31  & 14.37  & 13.89  \\
    \bottomrule
    \end{tabular}
    \vspace{-.1cm}
    \caption{\small \textbf{Comparisons of different caption types.} ``beam'' and ``nucleus'' denote beam search and nucleus sampling. ``Oracle'' denotes captions from COCO Caption. By default, we adopt COCO Caption fine-tuned BLIP model. \textsuperscript{*} denotes that we adopt the pre-trained BLIP model.}
    \vspace{-.2cm}
  \label{tab:caption_type}
\end{table}}

{\renewcommand{\arraystretch}{0.9}
\begin{table}[t]
  \small
  \setlength{\tabcolsep}{3.8pt}
  \centering
    \begin{tabular}{cc|ccc}
    \toprule
    \textbf{Dataset} & \textbf{Relation candidate} & \textbf{Rare} & \textbf{Non-Rare} & \textbf{Full} \\
    \midrule
    \midrule
    VG    & -     & 12.12  & 14.07  & 13.62  \\
    \midrule
    VG+COCO & BLIP  & \textbf{15.08}  & \textbf{15.10}  & \textbf{15.09}  \\
    VG+COCO & Selection from VG & 10.34  & 11.33  & 11.11  \\
    \bottomrule
    \end{tabular}
    \vspace{-.1cm}
    \caption{\small \textbf{Comparisons of methods to generate relation candidate sets.} We report zero-shot (NF) results on HICO-DET.}
    \vspace{-.2cm}
  \label{tab:necessity_of_BLIP}
\end{table}}

\begin{table}[t]
  % \small
  \scriptsize
  \setlength{\tabcolsep}{4pt}
  \centering
    \begin{tabular}{c|ccc}%{c|p{7.21em}p{7.54em}c}
    \toprule
          & \multicolumn{1}{c}{\textbf{VG}} & \multicolumn{1}{c}{\textbf{VG+COCO}} & \textbf{VG+COCO+O365} \\
    \midrule
    \midrule
    \textbf{ResNet-50} & 13.03 / 14.98 / 14.53 & 15.00 / 16.60 / 16.23 & 19.64 / 17.24 / 17.79 \\
    \textbf{Swin-T} & 13.01 / 16.06 / 15.35 & 17.13 / 18.74 / 18.37 & 21.24 / 19.47 / 19.87 \\
    \textbf{Swin-L} & 19.93 / 18.74 / 19.02 & 22.59 / 21.09 / 21.44 & 27.97 / 21.90 / 23.29 \\
    \bottomrule
    \end{tabular}
    \vspace{-.1cm}
    \caption{\small 
        \textbf{Model and dataset scaling experiments using RLIPv2-ParSeDA.}
        Results are evaluated on HICO-DET \textit{Rare}/\textit{Non-Rare}/\textit{Full} sets under zero-shot (NF) setting.
        }
    \vspace{-.2cm}
  \label{tab:model_dataset_scaling}
\end{table}

\begin{table}[t]
  \scriptsize
  % \small
  \setlength{\tabcolsep}{2pt}
  \centering
    \begin{tabular}{c|c|c|cc|cc|c}
    \toprule
    \multirow{2}[2]{*}{\textbf{Model}} & \multirow{2}[2]{*}{\textbf{Backbone}} & \multirow{2}[2]{*}{\textbf{Extra}} & \multirow{2}[2]{*}{${\rm mR}@50$} & \multirow{2}[2]{*}{${\rm R}@50$} & \multicolumn{2}{c|}{\textbf{wmAP}} & \multirow{2}[2]{*}{${\rm \bm{score}}_{wtd}$} \\
          &       &       &       &       & \textbf{rel} & \multicolumn{1}{c|}{\textbf{phr}} &  \\
    \midrule
    \midrule
    VCTree~\cite{tang2019vctree} & X101-F & -     & 33.91  & 74.08  & 34.16  & 33.11  & 40.21  \\
    G-RCNN~\cite{yang2018graph_rcnn} & X101-F & -     & 34.04  & 74.51  & 33.51  & 34.21  & 41.84  \\
    Motifs~\cite{zellers2018neuralmotif} & X101-F & -     & 32.68  & 71.63  & 29.91  & 31.59  & 38.93  \\
    Unbiased~\cite{tang2020unbiasedSG} & X101-F & -     & 35.47  & 69.30  & 30.74  & 32.80  & 39.27  \\
    GPS-Net~\cite{lin2020gps} & X101-F & -     & 35.26  & 74.81  & 32.85  & 33.98  & 41.69  \\
    RelDN~\cite{zhang2019RelDN} & R101  & -     & 36.80  & 72.75  & 29.87  & 30.42  & 38.67  \\
    BGNN~\cite{li2021BGNN}  & R101  & -     & 39.41  & 74.93  & 31.15  & 31.37  & 40.00  \\
    SGTR~\cite{Li2021SGTR}  & R101  & -     & 42.61  & 59.91  & 36.98  & 38.73  & 42.28  \\
    RelTR~\cite{cong2022RelTR} & R50   & -     & -     & 64.47  & 34.17  & 37.44  & 41.54  \\
    \midrule
    \rowcolor{mygray} RLIPv2-ParSeD & R50\textsuperscript{*}   & -  & 44.58  & 58.04  & 43.30  & 43.12  & 46.18  \\
    \rowcolor{mygray}       & R50\textsuperscript{*}   & -  & 44.88  & 60.20  & 44.73  & 43.36  & 47.28  \\
    \rowcolor{mygray}       & R50\textsuperscript{\dag}   & - & 45.59  & 61.15  & 45.71  & 43.73  & 48.01  \\
    \rowcolor{mygray}       & R50   &  \textbf{(i)} & 50.42  & 63.35  & 47.65  & 45.23  & 49.82  \\
    \rowcolor{mygray}       & R50   &  \textbf{(ii)} & \textbf{52.07}  & 64.53  & 49.14  & \textbf{46.14}  & 51.01  \\
    \rowcolor{mygray}       & R50   &  \textbf{(iii)} & 51.31  & \textbf{65.99}  & \textbf{49.54}  & 45.71  & \textbf{51.30} \\
    \rowcolor{mygray}       & Swin-T &  \textbf{(iii)} & \textbf{59.61} & \textbf{68.81}  & \textbf{52.70}  & \textbf{48.01}  & \textbf{54.05} \\
    % \multirow{-6}[1]{*}{RLIPv2-ParSeDA}
    \rowcolor{mygray}  \multirow{-7}[1]{*}{RLIPv2-ParSeDA}     & Swin-L   &  \textbf{(iii)} & \textbf{64.72}  & \textbf{72.49}  & \textbf{56.38}  & \textbf{50.70}  & \textbf{57.34} \\
    \bottomrule
    \end{tabular}
    \vspace{-.1cm}
    \caption{\small \textbf{Comparisons with previous methods on Open Images v6 SGG benchmark.} X101-F denote ResNeXt-101 FPN~\cite{xie2017resnext}. \textsuperscript{*} and \textsuperscript{\dag} denote ImageNet pretrained and COCO object detection pre-trained. ``\textbf{Extra}'' denotes extra relations adopted from \textbf{(i)} VG, \textbf{(ii)} VG+COCO and \textbf{(iii)} VG+COCO+O365.}
    \vspace{-.2cm}
  \label{tab:SOTA_SGG}
\end{table}

% CLIP prompt 介绍一下
\textbf{Comparisons of different relation tagging strategies.} 
We compare R-Tagger with other two methods:
\textbf{(i)} the greedy matching algorithm~\cite{zhong2021scene_graph_language};
\textbf{(ii)} the CLIP~\cite{radford2021CLIP} tagging method.
Specifically, to tag relations for a given SO region pair with a candidate relation text, the CLIP tagging method clips the minimum bounding box of the SO region pair, creates two prompts (``a photo of \{subject\} \{relation\} \{object\}'' and ``a photo of \{subject\} not interacting with \{object\}'' are adopted as positive and negative prompts.), and performs zero-shot prediction.
If the {\rm softmax} probability of the positive prompt is greater than a pre-defined threshold, we tag this relation text to this SO region pair.
(We traverse the threshold and find the optimal one for the CLIP tagging method, as detailed in the Appendix.)
We also ablate on the ``overlap'' prior~\cite{zhong2021scene_graph_language} to observe whether a given tagging method relies on strong prior knowledge to filter out false positive relations.
The results are shown in~\cref{tab:tagging_strategy}.
From this table, we conclude that greedy matching and CLIP tagging method generate a significant number of low-quality non-overlapped triplets.
Thus, the ``overlap'' prior is essential for them.
R-Tagger, however, suffers a slight performance drop when using the ``overlap'' prior, indicating that R-Tagger generates more reliable non-overlapped triplets.
Besides, although CLIP is pre-trained on a massive quantity of language-image pairs, it struggles in recognizing relations, which R-Tagger is expert in.

\textbf{Comparisons of different caption types.}
Previous experiments demonstrate that oracle COCO captions can benefit pre-training.
However, most datasets lack such high-quality captions.
Thus, BLIP provides an alternative for caption generation.
We conduct experiments on various caption types, as shown in~\cref{tab:caption_type}.
From this table, we can observe that:
\textbf{(i)} BLIP-generated captions achieve a decent performance compared to oracle captions, indicating the practicality of adopting generated captions for pseudo-labelling.
\textbf{(ii)} If we compare BLIP model with different parameters (COCO Caption fine-tuned or only pre-trained), we can see that the fine-tuned model generates better captions with more boost on the \textit{Rare} set ($12.31 \rightarrow 14.67$). 
We conjecture that the caption quality of the curated style dataset COCO Caption is better than pre-training captions harvested from the web. 
\textbf{(iii)} By adopting generated captions, we can increase the number of captions per image, thus diversifying the relations contained in captions and boost RLIPv2 ($11.52 \rightarrow 15.09$).
Besides, although datasets like Conceptual Caption~\cite{sharma2018CC3M} (CC3M) also provide captions, the quantity of captions (average one caption per image) is not enough to describe many relations in an image.
In comparison, our pipeline can work without caption annotation while performing better.
% pre-trained的web scale数据没有curated好
% 讲一下和之前文章的区别，只有一个caption，而且caption的质量不高。

{\renewcommand{\arraystretch}{0.9}
\begin{table}[t]
  % \small
  \scriptsize
  \setlength{\tabcolsep}{4pt}
  \centering
    \begin{tabular}{c|c|c|c}
    \toprule
    \textbf{Method} & \textbf{Backbone} & \textbf{UC-RF} & \textbf{UC-NF} \\
    \midrule
    \midrule
    VCL~\cite{hou2020VCL} & ResNet-50 & 10.06 / 24.28 / 21.43 & 16.22 / 18.52 / 18.06 \\
    ATL~\cite{hou2021ATL} & ResNet-50 & 9.18 / 24.67 / 21.57 & 18.25 / 18.78 / 18.67 \\
    FCL~\cite{hou2021FCL} & ResNet-50 & 13.16 / 24.23 / 22.01 & 18.66 / 19.55 / 19.37 \\
    GEN-VLKT~\cite{GEN_VLKT} & ResNet-50 & 21,36 / 32.91 / 30.56 & \textbf{25.05} / 23.38 / 23.71 \\
    RLIPv1-ParSeD~\cite{Yuan2022RLIP} &  ResNet-50 & 16.43 / 30.59 / 27.76 &  16.99 / 24.71 / 22.93 \\
    RLIPv1-ParSe~\cite{Yuan2022RLIP} & ResNet-50 & 19.19 / 33.35 / 30.52 & 20.27 / 27.67 / 26.19 \\
    \midrule
    % RLIPv2-ParSeDA & \textbf{21.45} / \textbf{35.85} / \textbf{32.97} & 22.81 / \textbf{29.52} / \textbf{28.18} \\
    RLIPv2-ParSeDA & ResNet-50 & \textbf{21.45} / \textbf{35.85} / \textbf{32.97} & 22.81 / \textbf{29.52} / \textbf{28.18} \\
    RLIPv2-ParSeDA & Swin-T & \textbf{26.95} / \textbf{39.92} / \textbf{37.32} & 21.07 / \textbf{35.07} / \textbf{32.27} \\
    RLIPv2-ParSeDA & Swin-L & \textbf{31.23} / \textbf{45.01} / \textbf{42.26} & 22.65 / \textbf{40.51} / \textbf{36.94} \\
    \bottomrule
    \end{tabular}
    \vspace{-.1cm}
    \caption{\small \textbf{Comparisons with methods on HICO-DET under UC-RF and UC-NF settings.} We adopt ResNet-50 as the backbone. Results are reported on \textit{Unseen}/\textit{Seen}/\textit{Full} sets.}
    \vspace{-.2cm}
  \label{tab:zero-shot_UC-NF_UC-RF}
\end{table}}

{\renewcommand{\arraystretch}{0.9}
\begin{table}[t]
  % \small
  \scriptsize
  \setlength{\tabcolsep}{4pt}
  \centering
    \begin{tabular}{c|c|c|c}
    \toprule
    \textbf{Method} & \textbf{Backbone} & \textbf{1\% Data} & \textbf{10\% Data} \\
    \midrule
    \midrule
    RLIPv1-ParSeD~\cite{Yuan2022RLIP} & ResNet-50 & 16.22 / 18.92 / 18.30 & 15.89 / 23.94 / 22.09 \\
    RLIPv1-ParSe~\cite{Yuan2022RLIP} & ResNet-50 & 17.47 / 18.76 / 18.46 & 20.16 / 23.32 / 22.59 \\
    \midrule
    RLIPv2-ParSeDA & ResNet-50 & \textbf{22.13} / \textbf{24.51} / \textbf{23.96} & \textbf{23.28} / \textbf{30.02} / \textbf{28.46} \\
    RLIPv2-ParSeDA & Swin-T & \textbf{24.26} / \textbf{28.92} / \textbf{27.85} & \textbf{28.31} / \textbf{32.93} / \textbf{31.87} \\
    RLIPv2-ParSeDA & Swin-L & \textbf{31.89} / \textbf{32.32} / \textbf{32.22} & \textbf{34.75} / \textbf{38.27} / \textbf{37.46} \\
    % RLIPv2-ParSeDA\textsuperscript{*} & \textbf{31.89} / \textbf{32.32} / \textbf{32.22} & \textbf{34.75} / \textbf{38.27} / \textbf{37.46} \\
    \bottomrule
    \end{tabular}
    \vspace{-.1cm}
    \caption{\small \textbf{Comparisons on HICO-DET under few-shot settings.} Results are reported on \textit{Rare}/\textit{Non-Rare}/\textit{Full} sets. 
    % ResNet-50 is adopted by default. \textsuperscript{*} denotes Swin-L is adopted.
    }
    \vspace{-.2cm}
  \label{tab:few-shot_transfer}
\end{table}}

{\renewcommand{\arraystretch}{0.9}
\begin{table*}[t]
  \centering
  % \scriptsize
  \small
  \setlength{\tabcolsep}{4pt}
    \begin{tabular}{c|c|c|cc|cc}
    \toprule
    \multirow{2}[2]{*}{\textbf{Model}} & \multirow{2}[2]{*}{\textbf{Backbone}} & \multirow{2}[2]{*}{\textbf{Extra Relations}} & \multicolumn{2}{c}{\textbf{HICO-DET}} & \multicolumn{2}{|c}{\textbf{V-COCO}} \\
          &       &       & \textbf{Zero-shot (NF)} & \textbf{Fully-finetuning} & ${\rm AP}_{role}^{\#1}$ & ${\rm AP}_{role}^{\#2}$ \\
    \midrule
    \midrule
    InteractNet~\cite{gkioxari2018InteractNet} & R50-FPN & -     & -     & 7.16 / 10.77 / 9.94 & 40.0  & - \\
    UnionDet~\cite{kim2020uniondet} & R50-FPN & -     & -     & 11.72 / 19.33 / 17.58 & 47.5  & 56.2  \\
    PPDM~\cite{liao2020ppdm}  & HG104 & -     & -     & 13.97 / 24.32 / 21.94 & -     & - \\
    HOTR~\cite{kim2021hotr}  & R50   & -     & -     & 17.34 / 27.42 / 25.10 & 55.2  & 64.4  \\
    QPIC~\cite{tamura2021qpic}  & R50   & -     & -     & 21.85 / 31.23 / 29.07 & 58.8  & 61.0  \\
    OCN~\cite{Yuan2022OCN}   & R50   & -     & -     & 25.56 / 32.51 / 30.91 & 64.2  & 66.3  \\
    CDN~\cite{zhang2021CDN}   & R50   & -     & -     & 27.39 / 32.64 / 31.44 & 61.7  & 63.8  \\
    % QAHOI~\cite{chen2021qahoi} & Swin-T & -     & -     & 22.44 / 30.27 / 28.47 & -     & - \\
    GEN-VLKT~\cite{GEN_VLKT} & R50   & -     & -     & 29.25 / 35.10 / 33.75 & 62.4  & 64.5  \\
    QAHOI~\cite{chen2021qahoi} & Swin-L\textsuperscript{*} & -     & -     & 29.80 / 37.56 / 35.78 & -     & - \\
    UniVRD~\cite{zhao2023UniVRD} & ViT-H/14\textsuperscript{\dag} & - & - & 31.65 / 39.99 / 38.07 & 65.8  &  66.9 \\
    RLIPv1-ParSeD~\cite{Yuan2022RLIP} & R50   & VG    & 11.20 / 14.73 / 13.92 & 24.67 / 32.50 / 30.70 & 61.7  & 63.8  \\
    RLIPv1-ParSe~\cite{Yuan2022RLIP} & R50   & VG    & 15.08 / 15.50 / 15.40 & 26.85 / 34.63 / 32.84 & 61.9  & 64.2  \\
    \midrule
    \rowcolor{mygray} RLIPv2-ParSeDA & R50   & VG    & 13.03 / 14.98 / 14.53 & 27.01 / 35.21 / 33.32 &   63.0    &  65.1 \\
    \rowcolor{mygray} RLIPv2-ParSeDA & R50   & VG+COCO & 15.00 / 16.60 / 16.23 & 27.89 / 35.27 / 33.57 &    64.5   &  66.7\\
    \rowcolor{mygray} RLIPv2-ParSeDA & R50   & VG+COCO+O365 & \textbf{19.64} / \textbf{17.24} / \textbf{17.79} & \textbf{29.61} / \textbf{37.10} / \textbf{35.38} &   \textbf{65.9}    & \textbf{68.0} \\
    \rowcolor{mygray} RLIPv2-ParSeDA & Swin-T & VG+COCO+O365 & \textbf{21.24} / \textbf{19.47} / \textbf{19.87} &  \textbf{33.66} / \textbf{40.07} / \textbf{38.60} &  \textbf{68.8}  &  \textbf{70.8} \\
    \rowcolor{mygray} RLIPv2-ParSeDA & Swin-L & VG+COCO+O365 &   \textbf{27.97} / \textbf{21.90} / \textbf{23.29}    &  \textbf{43.23} / \textbf{45.64} / \textbf{45.09}  &  \textbf{72.1}   & \textbf{74.1} \\
    \bottomrule
    \end{tabular}
    \vspace{-.1cm}
    \caption{\small \textbf{Comparisons with previous methods on HICO-DET and V-COCO.} Results on HICO-DET are reported on \textit{Rare}/\textit{Non-Rare}/\textit{Full} sets. R50 and HG denote ResNet-50~\cite{he2016resnet} and Hourglass~\cite{newell2016hourglass}. 
    * denotes the backbone is pre-trained with $384 \times 384$ resolution, while others use $224 \times 224$.
    \dag \ indicates the backbone is pre-trained using LiT~\cite{zhai2022LiT}, then fine-tuned on Objects365, COCO and HICO with the objective of object detection.}
    \vspace{-.2cm}
  \label{tab:SOTA_HOI}
\end{table*}}

% $\dag$ indicates that the backbone network was initially pre-trained using the LiT methodology~\cite{zhai2022LiT}. Subsequently, it was fine-tuned on the Objects365, COCO, and HICO datasets with the objective of object detection.

\textbf{The necessity of using captioners.}
By default, we use BLIP to generate relation texts. Another option is to query all possible SO pairs and select all possible relation texts from VG as it contains an enormous quantity of relations (36,515 kinds).
Then, we run R-Tagger with selected relation texts as $\bm{T}$ and all region pairs as $\bm{P}$.
The results are shown in~\cref{tab:necessity_of_BLIP}.
We can observe that selecting from VG generates low-quality candidates, harming performance.

\textbf{Model scaling and dataset scaling.}
Equipped with the labelling pipeline introduced above, we can scale RLIPv2 to larger models and datasets.
In~\cref{tab:model_dataset_scaling}, we adopt RLIPv2-ParSeDA as the base architecture and observe the benefit of scaling by zero-shot (NF) performance on HICO-DET.
In terms of data, adding COCO and Objects365 can both boost performance, and the benefit of adding data exhibits a ${\rm log}$ scaling trend~\cite{cherti2022reproducible_scaling_law}.
Models pre-trained with Objects365 consistently have better \textit{Rare} result, which we attribute to the distribution misalignment of Objects365 and HICO-DET~\cite{entezari2023role_of_pre-training_data}.
In terms of models, switching to stronger backbone models can improve the data efficiency at the cost of larger amounts of computation.
\hangjie{Regarding scaling experiments using RLIPv2-ParSeD, we present it in the Appendix.}

% Although scaling both models and datasets can boost performance, . 

% \vspace{-.2cm}
\subsection{Comparisons with State-of-the-Arts}
% \subsubsection{Scene Graph Generation}

\textbf{Scene graph generation.}
% In this subsection, 
We compare RLIPv2 series models with previous methods on Open Images v6 in~\cref{tab:SOTA_SGG}.
We also report mean Recall@50 (${\rm mR}@50$) to better show the usefulness of RLIPv2.
Our findings suggest that
\textbf{(i)} with the assistance of DDETR family models, RLIPv2 can serve as a strong baseline without any pre-training;
\textbf{(ii)} naive object detection pre-training can boost the performance to some extent ($47.28 \rightarrow 48.01$), while pre-training on VG can further boost the performance especially on ${\rm mR}@50$ ($45.59 \rightarrow 50.42$);
\textbf{(iii)} adding pseudo-labelled relation annotations in pre-training or switching to stronger backbones both contribute to better performance.
However, the boost of adding Objects365 is negligible.
We attribute this to the distribution discrepancy of Objects365 and Open Images v6.
% We can observe from the table that 
% \textbf{(i)} with the assistance of DDETR family models, RLIPv2 can serve as a very strong baseline without any form of pre-training;
% \textbf{(ii)} the naive object detection pre-training can boost the performance a bit ($47.28 \rightarrow 48.01$), while pre-training on VG can further boost the performance especially on ${\rm mR}@50$ ($45.59 \rightarrow 50.42$), which is consistent with conclusions from RLIPv1;
% \textbf{(iii)} adding pseudo-labelled relation annotations in pre-training or switching to stronger backbones both contributes to better performance.
% However, the boost of adding Objects365 is trivial.
% We attribute this to the distribution discrepancy of Objects365 and Open Images v6.

\textbf{HOI Detection under UC-NF and UC-RF settings.}
We report results in~\cref{tab:zero-shot_UC-NF_UC-RF} on unseen combinations (UC) under UC-RF and UC-NF settings following~\cite{hou2020VCL,hou2021ATL}.
We only fine-tune for 10 epochs under UC-NF setting.
Our method outperforms previous methods except on one metric.
We attribute this to the strong transferability of CLIP features that GEN-VLKT adopts.
\hangjie{Regarding experiments using RLIPv2-ParSeD, we present it in the Appendix.}

\textbf{Few-shot HOI Detection.} 
We follow~\cite{Yuan2022RLIP} to only fine-tune 10 epochs on partial data (1\% and 10\%), results of which are shown in~\cref{tab:few-shot_transfer}.
We can observe significant improvements upon all metrics by scaling up pre-training compared with previous methods.
This improves the practicality of RLIPv2 in low-data scenarios.
\hangjie{Regarding experiments using RLIPv2-ParSeD, we present it in the Appendix.}

\begin{figure}[t]
\centering
\vspace{-0.4cm}
\includegraphics[width=0.44\textwidth]{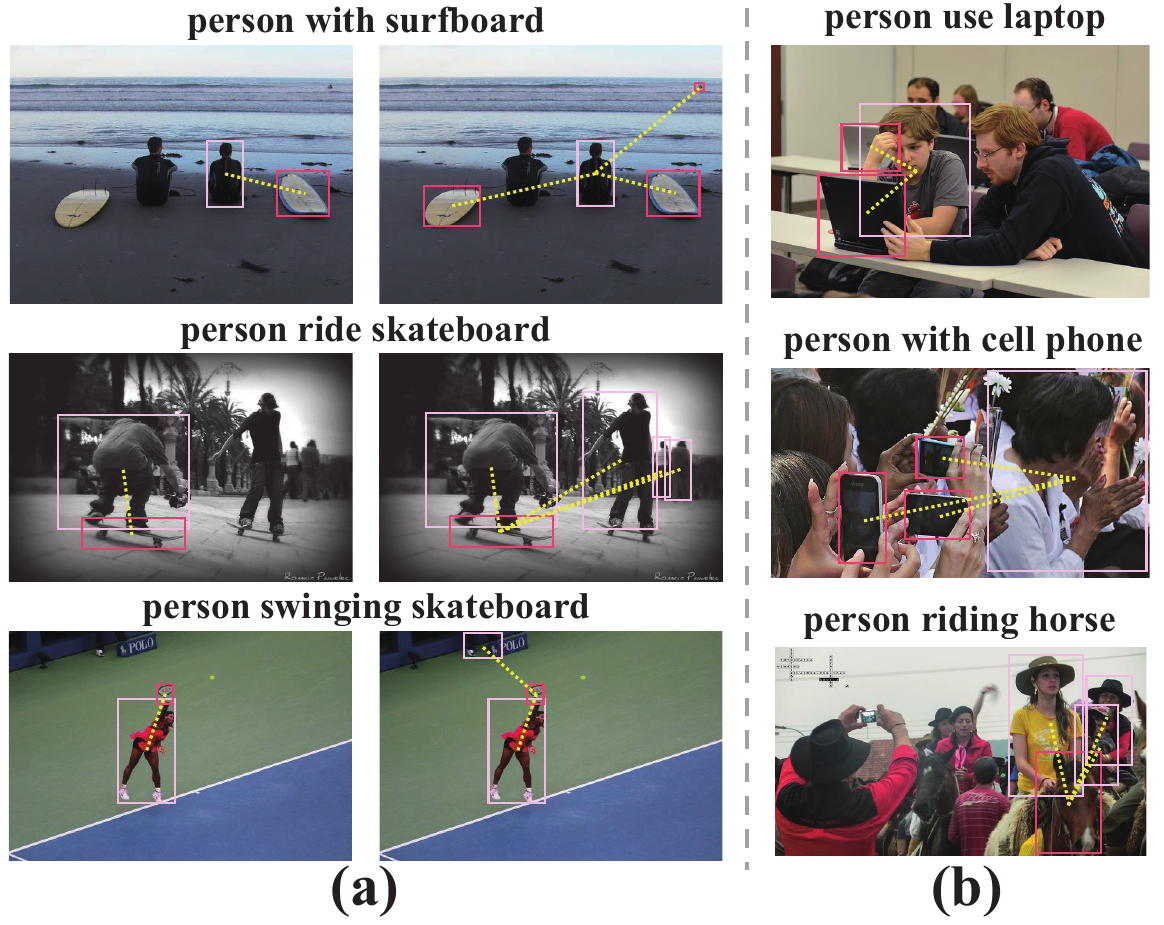}
% \vspace{-0.4cm} % for review version
\vspace{-0.2cm}
\caption{\small \textbf{(a) Visualization of pseudo-labelled relations on COCO}. Left column: R-Tagger; right column: CLIP tagging method. \textbf{(b) Visualization of failure cases of R-Tagger.}}
\vspace{-0.4cm}
\label{fig:pseudo-labels_visualization}
\end{figure}

% \subsubsection{HOI Detection}
\textbf{HOI detection under fully-finetuning and zero-shot (NF) settings.}
% In this subsection, 
We compare the performance of RLIPv2 series models with previous methods on HICO-DET and V-COCO in~\cref{tab:SOTA_HOI}.
We can observe from the table that
\textbf{(i)} dataset and model scaling can both boost the final performance on two datasets;
\textbf{(ii)} on HICO-DET, the benefit of pre-training is more prominent on zero-shot than fully-finetuning, especially on the \textit{Rare} set.
\hangjie{Regarding experiments using RLIPv2-ParSeD, we present it in the Appendix.}

\subsection{Qualitative Analysis}

\textbf{Comparisons of relation tagging methods.}
We visualize three examples to compare the quality of pseudo-labelled relations by R-Tagger and CLIP tagging method in~\cref{fig:pseudo-labels_visualization}(a).
Generally, CLIP is more object-centric and position-agnostic, and thus struggles in discriminating relations.
It tends to tag relations as long as the subject and object have strong co-occurrence priors.
However, R-Tagger tags more reasonable relations.

\textbf{Failure cases of R-Tagger.}
Recognizing relations is challenging, especially in complex scenes.
In~\cref{fig:pseudo-labels_visualization}(b), we present three examples of R-Tagger's failure cases.
In particular, we observe failure cases when the scene contains multiple similar subjects or objects.

% 前面结构那部分有一些表格还没加上去，包括实验结果。
% OI数据集的实验加进去
% Swin-L的scaling实验
% 最终的fine-tuning实验。
% hico加进去预训练
% fusion最后一层

% Scale up the dataset size or the model size?

% Validate R-Tagger with Oracle COCO captions
% \subsection{Ablation Study on ALIF}

% \subsection{Ablation Study on Relational Pseudo-labelling}

%% file: 06-conclusion.tex
\section{Conclusion}
In this paper, we propose RLIPv2, a fast converging model that enables the scaling of relational pre-training to larger-scale pseudo-labelled datasets.
Comprehensive experiments on HOI detection and scene graph generation under various settings demonstrate its effectiveness compared to previous methods.
We anticipate that our work can galvanize further research efforts to focus on relational reasoning, fostering advancements that yield tangible benefits not only to the research community but also to broader society and humanity.

% \noindent\textbf{Acknowledgements.} 
\subsection*{Acknowledgments}
We would like to extend our sincere gratitude to the anonymous reviewers for their invaluable feedback. 
Additionally, we appreciate the Fundamental Vision Intelligence Team of Alibaba DAMO Academy for their generous provision of essential computational resources.
This research received support from the National Natural Science Foundation of China under Grant No. 62173298 and was additionally backed by Alibaba Group via the Alibaba Research Intern Program.

% We would like to appreciate anonymous reviewers for their valuable feedback and members from Fundamental Vision Intelligence Team of Alibaba DAMO Academy for sharing computational resources. This work was supported in part by National Natural Science Foundation of China Grant No. 62173298 and by Alibaba Group through Alibaba Research Intern Program.

% \noindent\textbf{Acknowledgements.} Our gratitude extends to the anonymous reviewers, whose invaluable insights greatly enriched this work. We are also thankful to the Fundamental Vision Intelligence Team of Alibaba DAMO Academy for granting access to crucial computational resources. This research received support from the National Natural Science Foundation of China under Grant No. 62173298 and was additionally backed by Alibaba Group via the Alibaba Research Intern Program.

%% file: 08-appendix.tex
\appendix
\section{Appendix}
% I would probably move this paragraph to the Supplementary Material.

% \textbf{Revision of the main paper.}
In this Appendix, we first elaborate on the societal impact (\cref{app:societal_impact}), limitations (\cref{app:limitations}) and the use of datasets (\cref{app:datasets_used_in_this_work}) in RLIPv2.
% Next, we provide technical details of box embedding in Sec.~\textcolor{red}{4.2.2} of the main paper (\cref{sec:box_embedding}), cross attention mentioned in Eq.~(\textcolor{red}{2}) of the main paper (\cref{app:details_cross-attn}) and additional experiments (\cref{app:additional_experiments}).
Next, we provide technical details of the box embedding (\cref{sec:box_embedding}) described in Sec.~\textcolor{red}{4.2.2} of the main paper and cross attention modules (\cref{app:details_cross-attn}) described in Eq.~(\textcolor{red}{2}) of the main paper.
Finally, we present additional experiments (\cref{app:additional_experiments}) to further validate the effectiveness of our approach.

% Noise injection!!!!

% In this supplementary material, we first discuss the potential societal impact (Appendix A.1) and
% limitations (Appendix A.2) of our approach. Next, we provide further details on the architecture
% of RLIP-ParSeD (Appendix A.3), dataset pre-processing (Appendix A.4), phased pre-training (Ap-
% pendix A.5), attention analysis (Appendix A.6) and subject-object query pairing (Appendix A.7).
% Finally, we provide additional experiments and analysis (Appendix A.8) and discuss our use of
% datasets (Appendix A.9). Codes will be publicly available upon publication.

\section{Societal Impact} \label{app:societal_impact}
Our work RLIPv2 can potentially offer societal and commercial benefits.
From the data perspective, RLIPv2 proposes a relational pseudo-labelling pipeline that avoids time- and cost-intensive work and obtains reasonable relation labels.
From the pre-training perspective, RLIPv2 could perform more efficient pre-training and show prominent data efficiency that can potentially save computational cost.
Nonetheless, we acknowledge that HOI detection and SGG are dual-use, meaning that it can be used for beneficial and malicious purposes.
For example, the improved technologies can be applied to facilitate surveillance activities.
Moreover, due to the bias of the pre-training and fine-tuning datasets, our algorithms can not guarantee equal performance for all demographic groups.
Therefore, we emphasize that our pre-training and fine-tuning method is more of a proof-of-concept and requires rigorous evaluation and oversight when deploying for application.

\section{Limitations} \label{app:limitations}
% As introduced in the main paper, our framework requires an external captioner.
% This captioner generates captions that are used for parsing relations.
As mentioned in the main paper, our framework requires an external captioner to generate captions for relation parsing.
One limitation of our method is that the performance depends on the quality of the captions.
For instance, web-scale datasets for BLIP pre-training are usually noisy and lack diverse relation descriptions, \textit{e.g.}, some words might be excessively used but convey only ambiguous information like ``with'' and ``near''.
This is also confirmed by Tab.~\textcolor{red}{6} of the main paper, showing that fine-tuning on curated style
dataset like COCO Caption~\cite{chen2015COCO_Caption} is expected.
% The second limitation is that .
% 1. relying on external data. (This is not a good one.)
% 2. quality of pesudo-labels 
% 3. rely on the qulity of captions. (many captions do not have useful meanings like XX with XX)

\section{Datasets Used in This Work} \label{app:datasets_used_in_this_work}
\textbf{Licenses.}
We uses three datasets for pre-training and three datasets for downstream transfer in RLIPv2.
The following datasets are used, each governed by their license:
the Objects365~\cite{shao2019Objects365}, COCO~\cite{lin2014MSCOCO}, Visual Genome~\cite{krishna2017visualgenome} and Open Images v6~\cite{kuznetsova2020open_image} datasets are licensed under a Creative Commons Attribution 4.0 License;
the HICO-DET~\cite{chao2015hico,chao2018learningtodetectHOI} dataset is licensed under a CC0: Public Domain license;
the V-COCO~\cite{gupta2015VisualSemanticRole} dataset is licensed under an MIT license.

\textbf{Release of personally identifiable information/offensive content/consent.}
We affirm that no data will be disclosed as part of our research.
Our research relies on public domian datasets: Objects365~\cite{shao2019Objects365}, COCO~\cite{lin2014MSCOCO}, Visual Genome~\cite{krishna2017visualgenome}, Open Images v6~\cite{kuznetsova2020open_image}, HICO-DET~\cite{chao2015hico,chao2018learningtodetectHOI} and V-COCO~\cite{gupta2015VisualSemanticRole}, which we deem to pose a minimal risk of exposing personal information or offensive content. 
Regarding consent, we have not undertaken an independent inquiry beyond the scope of the original dataset releases.
% The sentence means that the authors of the paper have not done any investigation on their own beyond what was already done by the original creators of the datasets they used regarding whether the people whose data was included in the datasets gave their consent or not. This implies that there might be some ethical issues or risks involved in using the datasets.

% \section{Details about Noise Injection and Box Embedding}
% How to embed the position and label information.
% How to add noise
% As introduced in Sec.~\textcolor{red}{4.2.2} of the main paper, we manually add noise to the training stage of R-Tagger to ensure the training stability.
\section{Details about the Box Embedding} \label{sec:box_embedding}
% As mentioned in Sec.~\textcolor{red}{4.2.2} of the main paper, we embed the box positions and labels into queries to support for decoding.
As mentioned in Sec.~\textcolor{red}{4.2.2} of the main paper, we use box embeddings to encode labels and positions of the boxes $\hat{\bm{B}}_{s}, \hat{\bm{B}}_{o}$ into queries for decoding.
Specifically, regarding the label embeddings, we adopt the gradient-detached language features after ALIF.
To align the dimension of language features (\textit{i.e.}, $768$) with DDETR features (\textit{i.e.}, $256$), we apply linear projections to the language features.
Regarding the position embedding, we project the boxes $(x,y,w,h)$ into $256$ dimensions where $x, y$ are center coordinates and $w, h$ are width and height of the box.
Equipped with these two embeddings as queries, we can perform DDETR decoding~\cite{zhu2020deformableDETR}. 

{\renewcommand{\arraystretch}{0.9}
\begin{table}[t]
  \small
  \setlength{\tabcolsep}{8pt}
  \centering
    \begin{tabular}{cc|ccc}
    \toprule
          % &       &       &       &  \\
    \textbf{Threshold} & \textbf{Overlap} & \textbf{Rare} & \textbf{Non-Rare} & \textbf{Full} \\
    \midrule
    \midrule
    \multirow{2}[2]{*}{0.7} & \xmark     & 13.54  & 12.36  & 12.63  \\
          & \Checkmark     & 14.13  & \textbf{15.01}  & 14.81  \\
    \midrule
    \multirow{2}[2]{*}{0.8} & \xmark     & 12.66  & 12.76  & 12.74  \\
          & \Checkmark     & \textbf{14.63}  & 14.94  & \textbf{14.87}  \\
    \midrule
    \multirow{2}[2]{*}{0.9} & \xmark     & 12.00  & 12.38  & 12.49  \\
          & \Checkmark     & 13.28  & 14.35  & 14.10  \\
    \midrule
    \multirow{2}[2]{*}{0.95} & \xmark     & 11.66  & 12.08  & 11.98  \\
          & \Checkmark     & 13.69  & 14.37  & 14.21  \\
    \bottomrule
    \end{tabular}
    \vspace{-.2cm}
    \caption{\small \textbf{Parameter sensitivity analysis of the threshold for the CLIP tagging method.} ``Overlap'' denotes the ``overlap'' prior for SO pairs introduced in Sec.~\textcolor{red}{4.2.2}. We report zero-shot (NF) results pre-trained on VG and COCO.}
    \vspace{-.2cm}
  \label{tab:CLIP_threshold}
\end{table}}

{\renewcommand{\arraystretch}{0.9}
\begin{table}[t]
  \small
  \setlength{\tabcolsep}{14pt}
  \centering
    \begin{tabular}{c|ccc}
    \toprule
    \textbf{Threshold} $\eta$ & \textbf{Rare} & \textbf{Non-Rare} & \textbf{Full} \\
    \midrule
    \midrule
    -     & 12.12 & 14.07 & 13.62 \\ [-2pt]
    \midrule
    0.1   & 12.95 & 14.98 & 14.49 \\
    0.15  & 13.12 & 15.14 & 14.67 \\
    0.2   & \textbf{15.33} & \textbf{15.54} & \textbf{15.49} \\
    0.25  & 12.81 & 14.68 & 14.25 \\
    0.3   & 11.25 & 14.52 & 13.77 \\
    \bottomrule
    \end{tabular}
    \caption{\small \textbf{Parameter sensitivity analysis of $\eta$ for R-Tagger.} Zero-shot (NF) is reported after pre-training RLIPv2-ParSeD on VG and pseudo-labelled COCO. "-" denotes pre-training on VG.}
  \label{tab:R-Tagger_threshold}
\end{table}}

\section{Details about ${\rm Cross\text{-}attn}(\bm{C}^{(0)}, \bm{L}^{(0)})$} \label{app:details_cross-attn}
To perform cross attention as mentioned in Eq.~(\textcolor{red}{2}) of the main paper, we compute the attention scores of one modality with respect to the other modality, and then use scores to aggregate features from the other modality~\cite{dou2022FIBER,zhang2022glipv2,dou2022METER}.
Specifically, we follow the instantiation of the cross attention module from~\cite{li2021GLIP,zhang2022glipv2}. 
The calculation can be formulated as:
\begin{equation} \small
    \bm{C}^{(0,q)} = \bm{C}^{(0)} \bm{W}_{1}, \bm{L}^{(0,q)} = \bm{L}^{(0)} \bm{W}_{2}, Att = \frac{\bm{C}^{(0,q)} {(\bm{L}^{(0,q)})}^{T}}{\sqrt{d}}
\end{equation}
% \begin{equation} \small
%     \bm{C}^{(0,v)} = \bm{C}^{(0)} \bm{W}_{3}, \Tilde{\bm{C}}^{(0)} = {\rm softmax}(Att)\bm{C}^{(0,v)} \bm{W}_{4}
% \end{equation}
% \begin{equation} \small
%     \bm{L}^{(0,v)} = \bm{L}^{(0)} \bm{W}_{5}, \Tilde{\bm{L}}^{(0)} = {\rm softmax}({Att}^{T})\bm{L}^{(0,v)} \bm{W}_{6}
% \end{equation}
\begin{equation} \small
    \bm{L}^{(0,v)} = \bm{L}^{(0)} \bm{W}_{3}, \Tilde{\bm{C}}^{(0)} = {\rm softmax}(Att) \bm{L}^{(0,v)} \bm{W}_{4}
\end{equation}
\begin{equation} \small
    \bm{C}^{(0,v)} = \bm{C}^{(0)} \bm{W}_{5}, \Tilde{\bm{L}}^{(0)} = {\rm softmax}({Att}^{T}) \bm{C}^{(0,v)} \bm{W}_{6}
\end{equation}
where $\bm{W}_{1}, \bm{W}_{2}$ are trainable parameters for query embedding; $\bm{W}_{3}, \bm{W}_{5}$ are trainable parameters for value embedding; $\bm{W}_{4}, \bm{W}_{6}$ are trainable parameters for output embedding; $Att$ denotes attention scores; $d$ is embedding dimension, which is set to $256$ following~\cite{zhang2022glipv2,li2021GLIP}; the feature dimension of $\bm{C}^{(0)}$ and $\bm{L}^{(0)}$ are $256$ and $768$, respectively.
% use vision features from the last layer
Note that to reduce computation, we only perform cross attention on the flattened vision features from the last layer with the smallest scale.  

% \section{Details about Scene Graph Generation}

\begin{figure}[t]
\centering
\includegraphics[width=0.35\textwidth]{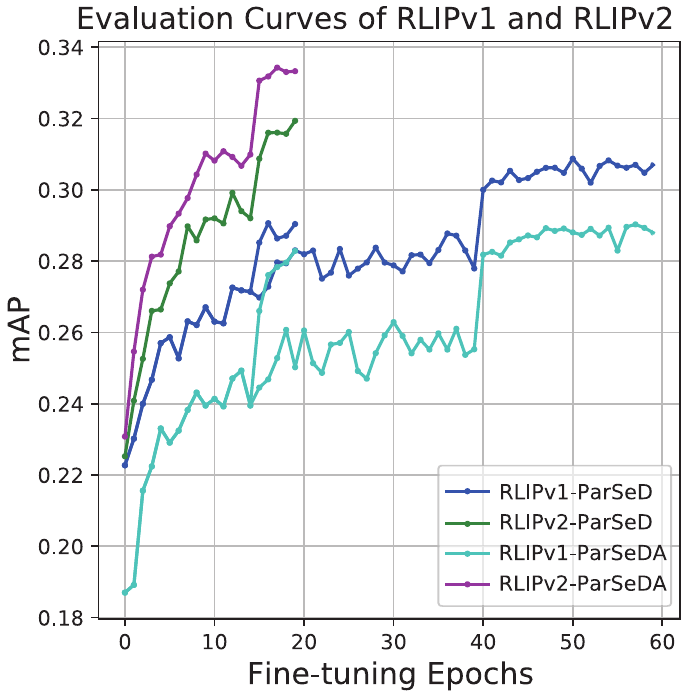}
\vspace{-0.2cm}
\caption{\small 
\textbf{Evaluation curve comparisons of RLIPv1 and RLIPv2 on HICO-DET.}
RLIPv1 and RLIPv2 are both pre-trained on VG.
By default, RLIPv1 is fine-tuned for 60 epochs following~\cite{Yuan2022RLIP} and RLIPv2 is fine-tuned for 20 epochs.
We also fine-tune RLIPv1 for 20 epochs for a fair comparison.
}
\vspace{-0.2cm}
\label{fig:evaluation_curves}
\end{figure}

{\renewcommand{\arraystretch}{0.9}
\begin{table}[t]
  \small
  \setlength{\tabcolsep}{4pt}
  \centering
    \begin{tabular}{cc|cc}
    \toprule
    \textbf{Model} & \textbf{Datasets} & \textbf{Epochs} & \textbf{Zero-shot (NF)} \\
    \midrule
    \midrule
    RLIP-ParSeDA & VG    & 50    & 11.34 / 14.56 / 13.82 \\
    RLIP-ParSeDA & VG+COCO & 50    & 15.05 / 15.55 / 15.44 \\
    \midrule
    RLIPv2-ParSeDA & VG    & 20    & 13.03 / 14.98 / 14.53 \\
    RLIPv2-ParSeDA & VG+COCO & 20    & 15.00 / 16.60 / 16.23 \\
    \bottomrule
    \end{tabular}
    \vspace{-.2cm}
    \caption{\small \textbf{Comparisons of scaling with different models.} We use ResNet-50 by default. Results are evaluated on HICO-DET under zero-shot (NF) setting.}
    \vspace{-.2cm}
  \label{tab:scaling_comparison}
\end{table}}

% {\renewcommand{\arraystretch}{0.9}
\begin{table}[t]
  \small
  \setlength{\tabcolsep}{3pt}
  \centering
    \begin{tabular}{cc|ccc}
    \toprule
    \textbf{Model} & \textbf{Pseudo-label Type} & \textbf{Rare} & \textbf{Non-Rare} & \textbf{Full} \\
    \midrule
    \midrule
    \multirow{2}[2]{*}{RLIPv2-ParSeDA} & R-Tagger & 15.00  & 16.60  & 16.23  \\
          & RLIPv2-ParSeD & 11.96  & 14.87  & 14.20  \\
    \bottomrule
    \end{tabular}
    \vspace{-.2cm}
    \caption{\small \textbf{Comparisons of different pseudo-label types.} RLIPv2-ParSeDA is pre-trained on VG and pseudo-labelled COCO. Results are evaluated on HICO-DET under zero-shot (NF) setting.}
    \vspace{-.2cm}
  \label{tab:pseudo-label_type_comparisons}
\end{table}

\begin{table*}[h]
  % \small
  % \vspace{-.3cm}
  % \scriptsize
  \setlength{\tabcolsep}{6pt}
  \centering
    \begin{tabular}{c|ccc}%{c|p{7.21em}p{7.54em}c}
    \toprule
          & \multicolumn{1}{c}{\textbf{VG}} & \multicolumn{1}{c}{\textbf{VG+COCO}} & \textbf{VG+COCO+O365} \\
    \midrule
    \midrule
    \textbf{ResNet-50} & 12.12 / 14.07 / 13.62 & 15.08 / 15.10 / 15.09 & 17.21 / 16.84 / 16.93 \\
    \textbf{Swin-T} & 12.17 / 15.01 / 14.36 & 14.89 / 16.70 / 16.28 &  20.34 / 18.27 / 18.75 \\
    \textbf{Swin-L} & 15.19 / 17.46 / 16.94 & 20.03 / 19.75 / 19.81 & 26.75 / 20.61 / 22.02  \\
    % \textbf{Swin-L} & 19.93 / 18.74 / 19.02 & 22.59 / 21.09 / 21.44 & 27.97 / 21.90 / 23.29 \\
    \bottomrule
    \end{tabular}
    % \vspace{-.2cm}
    \caption{\small 
        \textbf{Model and dataset scaling experiments using RLIPv2-ParSeD.}
        Results are evaluated on HICO-DET \textit{Rare}/\textit{Non-Rare}/\textit{Full} sets under zero-shot (NF) setting.
        }
    % \vspace{-.2cm}
    % \vspace{-.1cm}
  \label{tab:model_dataset_scaling_ParSeD}
\end{table*}

{\renewcommand{\arraystretch}{0.9}
\begin{table*}[h]
  % \small
  % \setlength{\tabcolsep}{4pt}
  \centering
    \begin{tabular}{c|c|cc}
    \toprule
    \textbf{Method} & \textbf{Backbone} & \textbf{UC-RF} & \textbf{UC-NF} \\
    \midrule
    \midrule
    VCL~\cite{hou2020VCL} & ResNet-50 & 10.06 / 24.28 / 21.43 & 16.22 / 18.52 / 18.06 \\
    ATL~\cite{hou2021ATL} & ResNet-50 & 9.18 / 24.67 / 21.57 & 18.25 / 18.78 / 18.67 \\
    FCL~\cite{hou2021FCL} & ResNet-50 & 13.16 / 24.23 / 22.01 & 18.66 / 19.55 / 19.37 \\
    GEN-VLKT~\cite{GEN_VLKT} & ResNet-50 & 21.36 / 32.91 / 30.56 & \textbf{25.05} / 23.38 / 23.71 \\
    RLIPv1-ParSeD~\cite{Yuan2022RLIP} &  ResNet-50 & 16.43 / 30.59 / 27.76 &  16.99 / 24.71 / 22.93 \\
    RLIP-ParSe~\cite{Yuan2022RLIP} & ResNet-50 & 19.19 / 33.35 / 30.52 & 20.27 / 27.67 / 26.19 \\
    \midrule
    RLIPv2-ParSeD &  ResNet-50 & \textbf{19.33} / \textbf{34.22} / \textbf{31.24} & 21.18 / \textbf{28.95} / \textbf{27.40} \\
    RLIPv2-ParSeD &  Swin-T & \textbf{23.80} / \textbf{38.23} / \textbf{35.34}  & 21.88 / \textbf{33.63} / \textbf{31.28} \\
    RLIPv2-ParSeD &  Swin-L & \textbf{30.98} / \textbf{43.67} / \textbf{41.13}  & 23.16 / \textbf{39.97} / \textbf{36.61} \\
    RLIPv2-ParSeDA & ResNet-50 & \textbf{21.45} / \textbf{35.85} / \textbf{32.97} & 22.81 / \textbf{29.52} / \textbf{28.18} \\
    RLIPv2-ParSeDA & Swin-T & \textbf{26.95} / \textbf{39.92} / \textbf{37.32} & 21.07 / \textbf{35.07} / \textbf{32.27} \\
    RLIPv2-ParSeDA & Swin-L & \textbf{31.23} / \textbf{45.01} / \textbf{42.26} & 22.65 / \textbf{40.51} / \textbf{36.94} \\
    \bottomrule
    \end{tabular}
    % \vspace{-.1cm}
    \caption{\small \textbf{Comparisons with methods on HICO-DET under UC-RF and UC-NF settings.} Results are reported on \textit{Unseen}/\textit{Seen}/\textit{Full} sets.}
    % \vspace{-.5cm}
  \label{tab:more_UC}
\end{table*}}

\section{Additional Experiments} \label{app:additional_experiments}
\textbf{The choice of the threshold for the CLIP tagging method.}
The CLIP tagging method requires a pre-defined threshold to tag relations as described in the Experiment section of the main paper (\textbf{Comparisons of different relation tagging strategies.}).
Therefore, to choose an optimal threshold, we traverse a range of values and evaluate the pre-training performance.
The results are presented in~\cref{tab:CLIP_threshold}.
From this table, we observe that utilizing the ``overlap'' prior can constantly outperform its naive counterpart without the ``overlap'' prior. 
Moreover, our analysis indicates that a threshold of $0.8$ leads to the best performance for the CLIP tagging method.
% The optimal threshold for the CLIP tagging method is $0.8$.

{\renewcommand{\arraystretch}{0.9}
\begin{table*}[t]
  \centering
    \begin{tabular}{cc|cc}
    \toprule
    \textbf{Method} & \textbf{Backbone} & \textbf{1\% Data} & \textbf{10\% Data} \\
    \midrule
    \midrule
    RLIP-ParSeD~\cite{Yuan2022RLIP} & ResNet-50 & 16.22 / 18.92 / 18.30 & 15.89 / 23.94 / 22.09 \\
    RLIP-ParSe~\cite{Yuan2022RLIP} & ResNet-50 & 17.47 / 18.76 / 18.46 & 20.16 / 23.32 / 22.59 \\
    \midrule
    RLIPv2-ParSeD & ResNet-50 & \textbf{19.87} / \textbf{24.04} / \textbf{23.08} & \textbf{21.51} / \textbf{27.84} / \textbf{26.38} \\
    RLIPv2-ParSeD & Swin-T & \textbf{26.37} / \textbf{27.29} / \textbf{27.08} & \textbf{27.85} / \textbf{31.41} / \textbf{30.59} \\
    RLIPv2-ParSeD & Swin-L & \textbf{30.49} / \textbf{30.80} / \textbf{30.73} & \textbf{33.90} / \textbf{35.93} / \textbf{35.46} \\
    RLIPv2-ParSeDA & ResNet-50 & \textbf{22.13} / \textbf{24.51} / \textbf{23.96} & \textbf{23.28} / \textbf{30.02} / \textbf{28.46} \\
    RLIPv2-ParSeDA & Swin-T & \textbf{24.26} / \textbf{28.92} / \textbf{27.85} & \textbf{28.31} / \textbf{32.93} / \textbf{31.87} \\
    RLIPv2-ParSeDA & Swin-L & \textbf{31.89} / \textbf{32.32} / \textbf{32.22} & \textbf{34.75} / \textbf{38.27} / \textbf{37.46} \\
    \bottomrule
    \end{tabular}
    % \vspace{-.1cm}
    \caption{\small \textbf{Comparisons with methods on HICO-DET under few-shot settings.} Results are reported on \textit{Rare}/\textit{Non-Rare}/\textit{Full} sets.}
    % \vspace{-.2cm}
  \label{tab:more_few-shot}
\end{table*}

{\renewcommand{\arraystretch}{0.9}
\begin{table*}[t]
  \centering
  % \scriptsize
  \small
  \setlength{\tabcolsep}{4pt}
    \begin{tabular}{c|c|c|cc|cc}
    \toprule
    \multirow{2}[2]{*}{\textbf{Model}} & \multirow{2}[2]{*}{\textbf{Backbone}} & \multirow{2}[2]{*}{\textbf{Extra Relations}} & \multicolumn{2}{c}{\textbf{HICO-DET}} & \multicolumn{2}{|c}{\textbf{V-COCO}} \\
          &       &       & \textbf{Zero-shot (NF)} & \textbf{Fully-finetuning} & ${\rm AP}_{role}^{\#1}$ & ${\rm AP}_{role}^{\#2}$ \\
    \midrule
    \midrule
    InteractNet~\cite{gkioxari2018InteractNet} & R50-FPN & -     & -     & 7.16 / 10.77 / 9.94 & 40.0  & - \\
    UnionDet~\cite{kim2020uniondet} & R50-FPN & -     & -     & 11.72 / 19.33 / 17.58 & 47.5  & 56.2  \\
    PPDM~\cite{liao2020ppdm}  & HG104 & -     & -     & 13.97 / 24.32 / 21.94 & -     & - \\
    HOTR~\cite{kim2021hotr}  & R50   & -     & -     & 17.34 / 27.42 / 25.10 & 55.2  & 64.4  \\
    QPIC~\cite{tamura2021qpic}  & R50   & -     & -     & 21.85 / 31.23 / 29.07 & 58.8  & 61.0  \\
    OCN~\cite{Yuan2022OCN}   & R50   & -     & -     & 25.56 / 32.51 / 30.91 & 64.2  & 66.3  \\
    CDN~\cite{zhang2021CDN}   & R50   & -     & -     & 27.39 / 32.64 / 31.44 & 61.7  & 63.8  \\
    % QAHOI~\cite{chen2021qahoi} & Swin-T & -     & -     & 22.44 / 30.27 / 28.47 & -     & - \\
    GEN-VLKT~\cite{GEN_VLKT} & R50   & -     & -     & 29.25 / 35.10 / 33.75 & 62.4  & 64.5  \\
    QAHOI~\cite{chen2021qahoi} & Swin-L\textsuperscript{*} & -     & -     & 29.80 / 37.56 / 35.78 & -     & - \\
    UniVRD~\cite{zhao2023UniVRD} & ViT-H/14\textsuperscript{\dag} & - & - & 31.65 / 39.99 / 38.07 & 65.8  &  66.9 \\
    RLIPv1-ParSeD~\cite{Yuan2022RLIP} & R50   & VG    & 11.20 / 14.73 / 13.92 & 24.67 / 32.50 / 30.70 & 61.7  & 63.8  \\
    RLIPv1-ParSe~\cite{Yuan2022RLIP} & R50   & VG    & 15.08 / 15.50 / 15.40 & 26.85 / 34.63 / 32.84 & 61.9  & 64.2  \\
    \midrule
    \rowcolor{mygray} RLIPv2-ParSeD & R50   & VG & 12.12 / 14.07 / 13.62 & 26.47 / 33.51 / 31.89 &  61.9   & 64.5 \\
    \rowcolor{mygray} RLIPv2-ParSeD & R50   & VG+COCO & 15.08 / 15.10 / 15.09 &  26.61 / 33.78 / 32.13 &  62.9   & 65.3 \\
    \rowcolor{mygray} RLIPv2-ParSeD & R50   & VG+COCO+O365 & \textbf{17.21} / \textbf{16.84} / \textbf{16.93} & \textbf{27.27} / \textbf{35.08} / \textbf{33.29} &  \textbf{63.8}    & \textbf{66.4} \\
    \rowcolor{mygray} RLIPv2-ParSeD & Swin-T   & VG+COCO+O365 & \textbf{20.34} / \textbf{18.27} / \textbf{18.75} &  \textbf{31.44} / \textbf{38.51} / \textbf{36.89} &  \textbf{66.6}   & \textbf{69.1} \\
    \rowcolor{mygray} RLIPv2-ParSeD & Swin-L   & VG+COCO+O365 & \textbf{26.75} / \textbf{20.61} / \textbf{22.02} & \textbf{42.76} / \textbf{44.67} / \textbf{44.23} &  \textbf{71.0}   & \textbf{73.2} \\
    \midrule
    \rowcolor{mygray} RLIPv2-ParSeDA & R50   & VG    & 13.03 / 14.98 / 14.53 & 27.01 / 35.21 / 33.32 &   63.0    &  65.1 \\
    \rowcolor{mygray} RLIPv2-ParSeDA & R50   & VG+COCO & 15.00 / 16.60 / 16.23 & 27.89 / 35.27 / 33.57 &    64.5   &  66.7\\
    \rowcolor{mygray} RLIPv2-ParSeDA & R50   & VG+COCO+O365 & \textbf{19.64} / \textbf{17.24} / \textbf{17.79} & \textbf{29.61} / \textbf{37.10} / \textbf{35.38} &   \textbf{65.9}    & \textbf{68.0} \\
    \rowcolor{mygray} RLIPv2-ParSeDA & Swin-T & VG+COCO+O365 & \textbf{21.24} / \textbf{19.47} / \textbf{19.87} &  \textbf{33.66} / \textbf{40.07} / \textbf{38.60} &  \textbf{68.8}  &  \textbf{70.8} \\
    \rowcolor{mygray} RLIPv2-ParSeDA & Swin-L & VG+COCO+O365 &   \textbf{27.97} / \textbf{21.90} / \textbf{23.29}    &  \textbf{43.23} / \textbf{45.64} / \textbf{45.09}  &  \textbf{72.1}   & \textbf{74.1} \\
    \bottomrule
    \end{tabular}
    % \vspace{-.2cm}
    \caption{\small \textbf{Comparisons with previous methods on HICO-DET and V-COCO.} Results on HICO-DET are reported on \textit{Rare}/\textit{Non-Rare}/\textit{Full} sets. R50 and HG denote ResNet-50~\cite{he2016resnet} and Hourglass~\cite{newell2016hourglass}. 
    * denotes the backbone is pre-trained with $384 \times 384$ resolution, while others use $224 \times 224$.
    \dag indicates the backbone is pre-trained using LiT~\cite{zhai2022LiT}, then fine-tuned on Objects365, COCO and HICO with the objective of object detection.}
    % \vspace{-.4cm}
  \label{tab:more_SOTA_HOI}
\end{table*}}

{\renewcommand{\arraystretch}{0.9}
\begin{table*}[h]
  \vspace{-.3cm}
  % \small
  % \scriptsize
  \setlength{\tabcolsep}{6pt}
  \centering
    \begin{tabular}{c|cc|ccc}
    \toprule
    \textbf{Caption type} & $N_{Cap}$ & $N_{Unique}$ & \textbf{Rare} & \textbf{Non-Rare} & \textbf{Full}  \\
    \midrule
    % - (baseline: \textit{w/o} COCO relations)
    - (baseline: \textit{w/o} captions) & - & - & 12.12 & 14.07 & 13.62 \\
    \midrule
    BLIP (beam) & 1  & 1   & 9.86  & 12.02  & 11.52  \\
    BLIP (nucleus) & 10  & 9.97  & 15.08 & 15.10 & 15.09  \\
    \midrule
    BLIP-2 (beam) & 1  & 1   & 9.98 & 12.23 & 11.72  \\
    BLIP-2 (nucleus) & 10  & 3.26  & 11.76 & 12.85 & 12.60  \\
    BLIP + BLIP-2 (nucleus) & 20 & 13.18 & 14.74 & 15.52 & \textbf{15.34}  \\
    \midrule
    BLIP (dense captions, beam) & 28.63 & 10.40 & 14.25 & 15.14 & 14.94 \\
    \bottomrule
    \end{tabular}
    \vspace{-.1cm}
    \caption{\small \textbf{Comparisons of different captioners.} 
    ``beam'' and ``nucleus'' denote beam search and nucleus sampling. 
    $N_{Unique}$ denotes the number of unique captions after deduplication.
    By default, we adopt COCO Caption fine-tuned BLIP and BLIP-2 model.
    }
    \vspace{-.2cm}
    % \vspace{-.4cm}
  \label{tab:diverse_captioners}
\end{table*}}

\textbf{The choice of threshold $\eta$ for R-Tagger.}
To choose pseudo-labelled triplets for pre-training, we traverse a range of $\eta$ values for R-Tagger to select triplets with relation confidence higher than this threshold.
% To generate an authentic relation candidate set for COCO, we adopt oracle captions from \textbf{COCO Caption}~\cite{chen2015COCO_Caption}, where each image has an average of $N_{Cap}=5$ captions.
By default, we adopt oracle captions from \textbf{COCO Caption}~\cite{chen2015COCO_Caption}, which provides an average of $N_{Cap}=5$ captions per image.
We tag pseudo-labels on COCO and pre-train RLIPv2-ParSeD on VG and pseudo-labelled COCO.
The results are presented in~\cref{tab:R-Tagger_threshold}, which indicates that the optimal value for $\eta$ is $0.2$.
It is worth noting that all experiments in this table are initialized with COCO object detection parameters.
Therefore, the highest performance boost ($13.62 \rightarrow 15.49$) is obtained by the additional tagged relations from COCO, rather than by the inclusion of an additional COCO dataset.

% \textbf{About R-Tagger.}
% An intuitive approach is executing the pre-trained RLIPv2 model to perform SGG on object detection datasets (e.g. COCO and Objects365) to obtain pseudo triplets.
% However, this process bears two drawbacks:
% \textbf{(i)} specifying a suitable set of relation candidates for a given image is challenging;
% \textbf{(ii)} specifying SO region pairs that might have valid relations for a given image is challenging;
% \textbf{(iii)} during the execution of RLIPv2, the informative object annotations from object detection datasets fail to be utilized, thus degrading the quality of pseudo-labels.
% Aiming to solve these, we propose a framework that utilizes a BLIP captioner~\cite{li2022blip} to generate open-vocabulary relation candidate sets for images and a designed Relation Tagger (R-Tagger) to tag relations provided object annotations and relation candidate sets.

% \textbf{About VG selection method.}
% \textcolor{red}{Can we delete it the following paragraph?}
% We also compare with the baseline of selecting the candidate set from VG annotations (select the relation texts if the subject and object texts are matched).
% However, the quality of such a selection method largely trails the proposed method, as shown in~\cref{tab:necessity_of_BLIP}.

\textbf{Evaluation curve comparisons.}
As detailed in Tab.~\textcolor{red}{3} of the main paper, we show the comparisons of RLIPv1 and RLIPv2 concerning pre-training and fine-tuning.
To further validate the effectiveness of ALIF, we compare their fine-tuning evaluation curves on HICO-DET in~\cref{fig:evaluation_curves}.
% We compare the evaluation curves of RLIPv1 and RLIPv2 in~\cref{fig:evaluation_curves}.
As can be observed from the figure, RLIPv2 can converge much faster than its RLIPv1 counterparts.

% Explain why RLIP-ParSeDA is not as good as RLIPv2-ParSeDA.
% %% DAB-DDETR在dedicated transformer以后预测位置能力就没有那么好了。
% We can also observe that RLIP-ParSeDA, which is modified from DAB-DDETR~\cite{liu2022DABDETR}, trails RLIP-ParSeD, which is modified from DDETR~\cite{zhu2020deformableDETR}.
% We attribute this to the dedicated attention layers used in RLIPv1 and the dynamically updated anchor boxes in DAB-DDETR.
% DAB-DDETR is designed to use dedicated channels to represent boxes. 

\textbf{Dataset scaling using RLIP.}
% One contribution of our paper is that RLIPv2 can facilitate fast scaling due to its convergence speed.
This paper introduces RLIPv2 as a novel method that facilitates scaling due to its convergence speed.
Another alternative is to adopt RLIPv1 to scale up relational pre-training.
In~\cref{tab:scaling_comparison}, we aim to compare the effect of scaling using RLIPv1 and RLIPv2.
We can observe that by performing earlier and deeper gated fusion, ALIF gains slightly better performance boost by costing $0.4\times$ pre-training time.
Therefore, RLIPv2 is a more efficient and scalable approach for scaling up relational pre-training.

% \textbf{Assigning relation labels via scene graph generation.}
\textbf{Comparisons of pseudo-label types.}
To validate the effectiveness of R-Tagger that utilizes groundtruth object annotations as input, we compare R-Tagger with another baseline that generates pseudo-labels by applying the pre-trained RLIPv2 model to perform SGG on object detection datasets as mentioned in Sec.~\textcolor{red}{4.2.2}.
% to obtain pseudo-triplets
% This paper proposes R-Tagger to assign relation texts to region pairs, while another alternative is to apply the pre-trained RLIPv2 model to perform SGG on object detection datasets to obtain pseudo-triplets by selection as mentioned in Sec.~\textcolor{red}{4.2.2}.
% We use an identical selection threshold to R-Tagger, \textit{i.e.}, $0.2$.
To leverage the annotated object boxes for the new baseline, we match the generated boxes with the annotated ones.
If a generated box is matched with one given annotated box (IoU $>$ 0.5), we substitute the generated box with the annotated box to ensure the accuracy of the box position.
To seek a fair comparison, we use the pre-trained RLIPv2-ParSeD (ResNet-50) to generated pseudo-triplets, which is identical to the base model of R-Tagger.
To select triplets, we use an identical selection threshold to R-Tagger, \textit{i.e.}, $0.2$.
Then, we pre-train on VG and pseudo-labelled COCO datasets to compare the quality of the pseudo-labels in~\cref{tab:pseudo-label_type_comparisons}.
As can be observed from the table, by utilizing groundtruth box information when inferring relations, R-Tagger can generate more authentic pseudo-labels, thus benefiting relational pre-training.

\textbf{Model scaling and dataset scaling using RLIPv2-ParSeD.}
In addition to scaling experiments using RLIPv2-ParSeDA in the main paper, we also present the model and dataset scaling experiments using RLIPv2-ParSeD in~\cref{tab:model_dataset_scaling_ParSeD}.
In terms of data, adding COCO and Objects365 can both boost performance, and the benefit of adding data exhibits a ${\rm log}$ scaling trend~\cite{cherti2022reproducible_scaling_law}.
Models pre-trained with Objects365 consistently have better \textit{Rare} result, which we attribute to the distribution misalignment of Objects365 and HICO-DET~\cite{entezari2023role_of_pre-training_data}.
In terms of models, switching to stronger backbone models can improve the data efficiency at the cost of larger amounts of computation.

\textbf{More results using RLIPv2-ParSeD on HICO-DET under UC-NF and UC-RF settings.}
We provide more results under UC-NF and UC-RF settings using RLIPv2-ParSeD in addition to RLIPv2-ParSeDA in~\cref{tab:more_UC}.
It is worth noting that UC-RF denotes 120 unseen combinations (UC) among 600 combinations are selected by a rare-first (RF) order, while UC-NF denotes that 120 unseen combinations among 600 combinations are selected by a non-rare first (NF) order.
We can observe that 
\textbf{(i)} under the UC-RF setting, switching to stronger backbones improves the performance of all metrics;
\textbf{(ii)} under the UC-NF setting, switching to stronger backbones enhances all metrics except the metric on the \textit{Unseen} set.
We attribute this to the significant object distribution misalignment between \textit{Seen} and \textit{Unseen} sets.
% 因为把non-rare中的triplet删掉以后，那些object标注在剩下的480类中在也是很少的，过少的数据会造成这种现象。
% 另一种极端情况是，当我们使用unseen object setting的时候，最终，由于catastrophic forgetting，unseen的性能会接近0。

\textbf{More results on few-shot HOI detection.}
% We provide more results under few-shot settings using ResNet-50~\cite{he2016resnet}, Swin-T~\cite{liu2021swin} and Swin-L~\cite{liu2021swin} in~\cref{tab:more_few-shot}.
We provide more results under few-shot settings using RLIPv2-ParSeD in addition to RLIPv2-ParSeDA in~\cref{tab:more_few-shot}.
We can observe that RLIPv2 exhibits remarkable data efficiency by scaling up pre-training.
Notably, the largest pre-trained model obtains 32.22mAP when fine-tuned on 1\% data, which outperforms many methods that fine-tune on 100\% data.

\textbf{More results under fully-finetuning and zero-shot (NF) settings on HOI detection.}
% We compare the performance of RLIPv2 series models with previous methods on HICO-DET and V-COCO in~\cref{tab:more_SOTA_HOI}.
We present more results using RLIPv2-ParSeD on HICO-DET and V-COCO in~\cref{tab:more_SOTA_HOI}.
% We can observe from the table that
We draw similar conclusions from this table that
\textbf{(i)} dataset and model scaling can both boost the final performance on two datasets;
\textbf{(ii)} on HICO-DET, the benefit of pre-training is more prominent on zero-shot than fully-finetuning, especially on the \textit{Rare} set.

\textbf{The effect of using diverse captioning models.}
To comprehensively study the effect of using other captioners, we adopt the more advanced BLIP-2~\cite{li2023blip-2} to implement our method.
The results are shown in~\cref{tab:diverse_captioners}.
The results indicate that BLIP-2 (beam) slightly outperforms BLIP (beam), but BLIP-2 (nucleus) trails BLIP (nucleus).
We attribute this to the low diversity of BLIP-2 captions as BLIP-2 generates more deterministic captions.
To verify this hypothesis, we combine captions from BLIP and BLIP-2, obtaining better generalization performance.

\textbf{The effect of dense captioning on pairwise union regions.}
% We experiment generating accurate captions using BLIP for pairwise union regions, as shown in the above table.
We show the result in the final row of the above table.
Dense captioning on pairwise union regions can improve over the baseline (the first row), offering an alternative to improve the caption diversity in addition to utilizing nucleus sampling.
Thus, the potential for developing more advanced captioning schemes to bolster the quality of pseudo-labels holds considerable promise.
% We leave it for future work.
We mark this endeavor as a research focus for one of our future works.